\pdfoutput=1

\documentclass[11pt]{article}

\usepackage[]{ACL2023}

\usepackage{times}
\usepackage{latexsym}
\usepackage{graphicx}
\usepackage[T1]{fontenc}
\usepackage{makecell}
\usepackage[utf8]{inputenc}
\usepackage{adjustbox}
\usepackage{graphicx}
\usepackage{microtype}
\usepackage{amssymb}
\usepackage{inconsolata}
\usepackage{booktabs}
\usepackage{multirow}
\usepackage{graphicx}

\usepackage{subcaption}
\title{Zero-Shot Cross-Domain Dialogue State Tracking via Dual Low-Rank Adaptation}

\author{
    Xiang Luo, Zhiwen Tang\thanks{\;\;Corresponding Authors}, Jin Wang{$^\ast$} and Xuejie Zhang\\
    School of Information Science and Engineering \\
    Yunnan University \\
    Kunming, China\\
    luoxiang@mail.ynu.edu.cn, \{zhiwen.tang, wangjin, xjzhang\}@ynu.edu.cn\\
    }

\begin{document}
\maketitle
\begin{abstract}

Zero-shot dialogue state tracking (DST) seeks to enable dialogue systems to transition to unfamiliar domains without manual annotation or extensive retraining. Prior research has approached this objective by embedding prompts into language models (LMs). Common methodologies include integrating prompts at the input layer or introducing learnable variables at each transformer layer. Nonetheless, each strategy exhibits inherent limitations. Prompts integrated at the input layer risk underutilization, with their impact potentially diminishing across successive transformer layers. Conversely, the addition of learnable variables to each layer can complicate the training process and increase inference latency. To tackle the issues mentioned above, this paper proposes Dual Low-Rank Adaptation (DualLoRA), a plug-and-play architecture designed for zero-shot DST. DualLoRA incorporates two distinct Low-Rank Adaptation (LoRA) components, targeting both dialogue context processing and prompt optimization, to ensure the comprehensive influence of prompts throughout the transformer model layers. This is achieved without incurring additional inference latency, showcasing an efficient integration into existing architectures. Through rigorous evaluation on the MultiWOZ and SGD datasets, DualLoRA demonstrates notable improvements across multiple domains, outperforming traditional baseline methods in zero-shot settings. Our code is accessible at: \url{https://github.com/suntea233/DualLoRA}.

\end{abstract}

\begin{figure}[!t]
    \centering
    \includegraphics[width=0.45\textwidth]{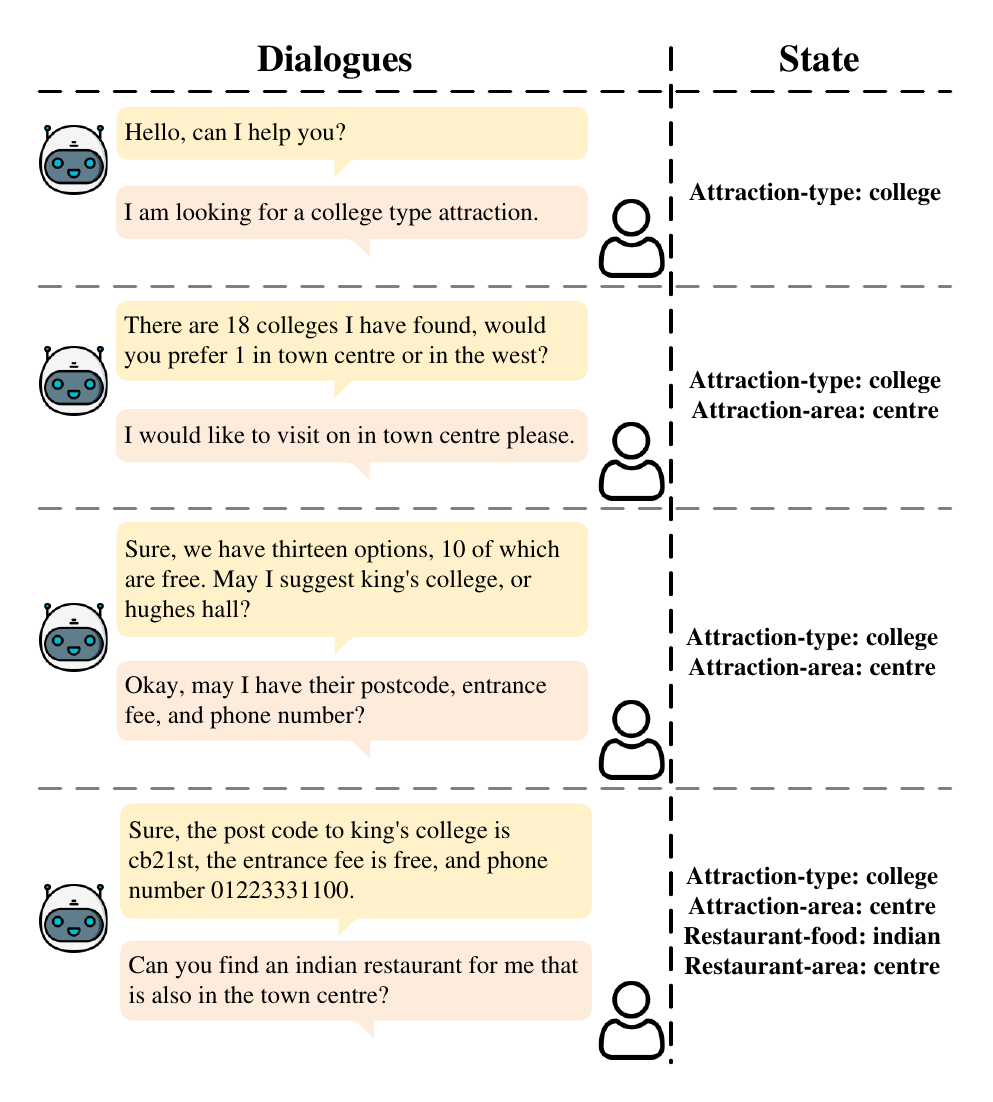}
    \caption{Example of a multi-domain dialogue from MultiWOZ dataset.}
    \label{fig: illustration}
\end{figure}

\section{Introduction}


Task-oriented dialogue (TOD) systems, designed for specific tasks like restaurant reservations or travel bookings, rely on dialogue state tracking (DST) to interpret user intents as slot-value pairs across dialogue turns, essential for managing multi-domain conversations \cite{huang_challenges_2020,luo-etal-2024-duetsim-building}. Figure \ref{fig: illustration} shows the change of dialogue states in a multi-turn dialogue crossing multiple domains.  An ideal system seamlessly transitions to new domains with minimal training, overcoming the resource-intensive process of data collection and annotation for new domains. Zero-shot learning emerges as a solution, enabling TOD systems to adapt to unseen domains by leveraging existing knowledge, thus bypassing the need for extensive domain-specific data.




\begin{figure*}[t]
    \centering
    \begin{subfigure}{0.28\textwidth}
        \includegraphics[width=\linewidth]{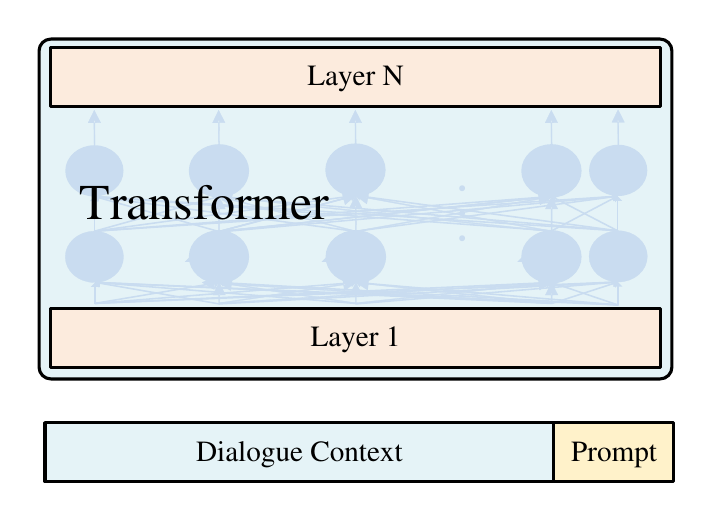}
        \caption{Type I: Incorporating prompts in the input layer.}
        \label{fig:single_llm1}
    \end{subfigure}
    \begin{subfigure}{0.28\textwidth}
        \includegraphics[width=\linewidth]{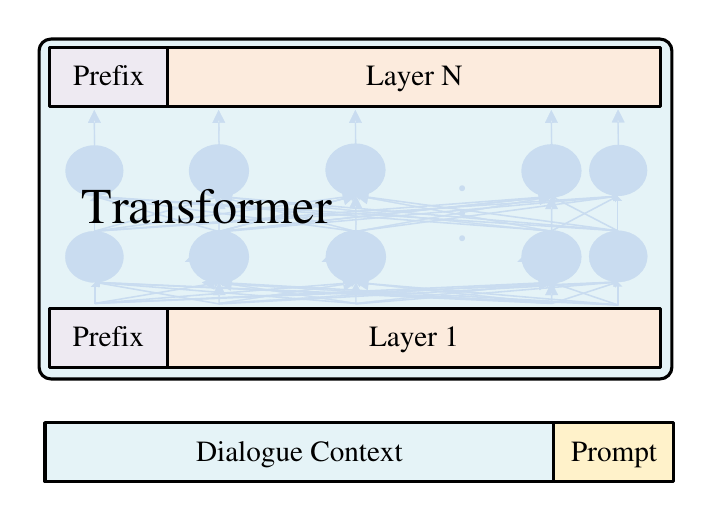}
        \caption{Type II: Adding prefixes in every layer.}
        \label{fig:single_llm2}
    \end{subfigure}
    \begin{subfigure}{0.33\textwidth}
        \includegraphics[width=\linewidth]{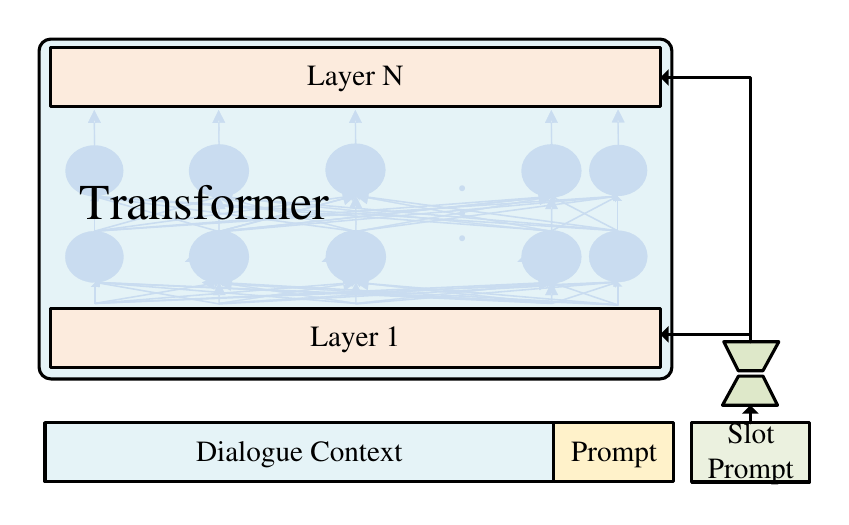}
        \caption{DualLoRA: Encoding prompts with a separate LoRA.}
        \label{fig:dual_llm}
    \end{subfigure}
    \caption{Different Architectures for Zero-Shot Dialogue State Tracking.}
    \label{fig: different generative approaches}
\end{figure*}

\begin{figure}[!t]
\centering


\begin{subfigure}{0.22\textwidth}
        \includegraphics[width=\linewidth]{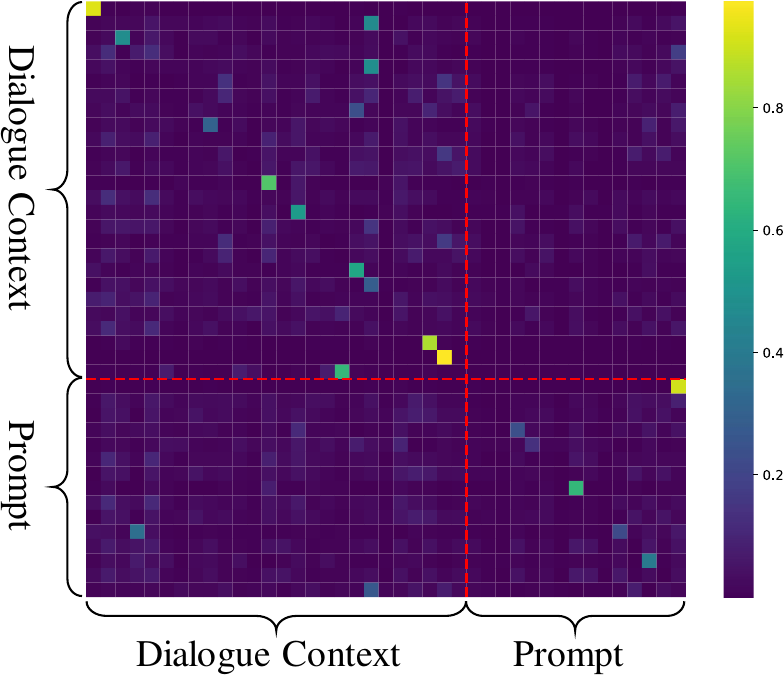}
        \caption{Attention map in the first encoder layer.}
\end{subfigure}
\begin{subfigure}{0.22\textwidth}
        \includegraphics[width=\linewidth]{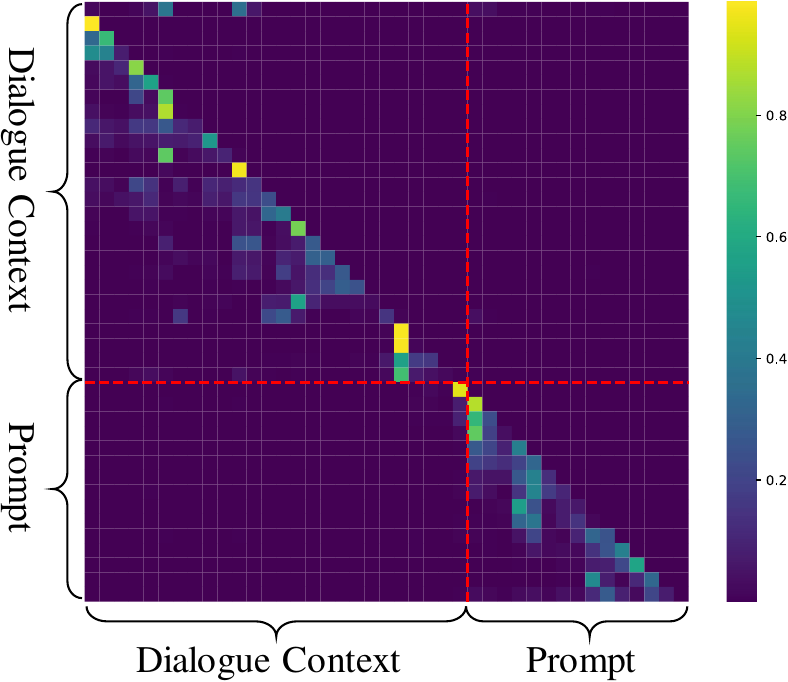}
        \caption{Attention map in the last encoder layer.}
\end{subfigure}

\caption{T5DST \cite{lin_leveraging_2021} attention map between dialogue context and prompts in the first layer and last layer of the encoder. The last encoder layer allocates less attention to prompt tokens.}
\label{fig: attn_map}
\end{figure}


The advancement in language models (LMs) \cite{zheng-etal-2024-enhancing-semantics} has significantly enhanced the exploration of zero-shot learning approaches within the context of dialogue state tracking tasks. Adaptation to novel domains is frequently achieved through the incorporation of prompts into transformer-based models \cite{lin_leveraging_2021}. The current method can be classified into two categories.

The first type of methods incorporate prompts, e.g. slot description, in the first transformer layer \cite{lin_leveraging_2021,lee_dialogue_2021,lin_zero-shot_2021,qixiang_exploiting_2022}. A common practice within this approach involves the concatenation of prompts subsequent to the input, as shown in Fig \ref{fig:single_llm1}. Despite the straightforward nature of their implementation, these strategies exhibit suboptimal utilization of prompts. This refers to the phenomenon where the influence of prompts on the model's output progressively diminishes with each additional transformer layer. 
Figure \ref{fig: attn_map} shows an example, where deep transformer layers, compared with shallow transformer layers, exhibit a reduced allocation of attention to other tokens within the dialogue context or prompt, in the token encoding process.  This issue is further exacerbated in the context of dialogue state tracking, where the dialogue's progression causes a rapid escalation in the context length — often extending to 300-400 tokens or more, in stark contrast to the prompts' modest length of 10-20 tokens. Such a diminishing effect of prompts can severely impair the system's performance in zero-shot learning scenarios, where the ability to discern and leverage domain-specific knowledge becomes critically important.

The second category of methodologies incorporates a learnable vector, initialized by prompts, into each layer of the transformer architecture \cite{aksu_prompter_2023}, as shown in Fig \ref{fig:single_llm2}. This technique facilitates the model's capacity to consistently engage with the prompt information across all layers, ensuring the prompt's considerations are integrated throughout the encoding process. However, these methods present certain limitations. Direct concatenation of prompts at each layer might introduce extraneous noise, adversely affecting the efficacy of model training. Furthermore, this method of concatenation can result in augmented inference latency of the model. Consequently, devising a strategy to effectively utilize prompts without impeding the standard input-output dynamics of the model and without exacerbating inference time emerges as a critical challenge in the field.


To address the aforementioned challenges, we propose a novel plug-and-play framework designed for zero-shot dialogue state tracking, termed Dual Low-Rank Adaptation (DualLoRA). This architecture is characterized by the integration of two Low-Rank Adaptation (LoRA) components: one dedicated to processing dialogue context and the other to refining prompts, as shown in Fig \ref{fig:dual_llm}. DualLoRA operates by receiving prompts and applying modifications within the attention layers of the transformer model, subsequently integrating this modified output with the transformer's original output. Through the implementation of this framework across each layer, DualLoRA ensures that the influence of the prompts is perpetuated throughout the entire depth of the model. Notably, this architecture does not incur additional latency during the inference phase, maintaining efficiency in model performance. Furthermore, DualLoRA requires few parameter adjustments and demonstrates exceptional transferability across different domains. 


Comprehensive experimental evaluations were performed utilizing the MultiWOZ and SGD datasets to assess the efficacy of the proposed methodology. Comparative analyses reveal that our approach yields enhanced Joint Goal Accuracy (JGA) metrics across various domains within both the MultiWOZ and SGD datasets, relative to established baseline methodologies.

The structure of this paper is organized as follows: Section 2 offers an overview of the literature pertinent to this research area. Section 3 delineates a detailed exposition of our proposed model. Section 4 presents a detailed account of the experimental outcomes and the specifics of the implementation. Section 5 encapsulates the conclusions derived from this study.

\section{Related Work}
\subsection{Dialogue State Tracking}
Recent works on dialogue state tracking have mainly approached the task in two ways. One approach treats the task as a classification task, utilizing an encoder to classify or extract dialogue states. Such as \cite{rastogi_scalable_2017,zhong_global-locally_2018,ma_end--end_2019,lee_sumbt_2019,chao_bert-dst_2019,gao_machine_2020,heck_trippy_2020,dai_preview_2021,balaraman_domain-aware_2021}, which employs BERT or LSTM or GRU or RNN as the context encoder for the dialogue, extracting slot values from the semantic context. However, since these methods are trained on a large amount of annotated data across multiple domains, the performance tends to be poor when dealing with unseen domains, indicating limited generalization. The other approaches treat the task as a generative task, employing a Seq2Seq approach, such as the T5 \cite{raffel_exploring_2020} and GPT \cite{radford_language_2019} model. Such as \cite{rastogi_scaling_2019,wu_transferable_2019,lin_leveraging_2021,lin_zero-shot_2021,li_zero-shot_2021,lee_dialogue_2021,qixiang_exploiting_2022,yu_knowledge-grounded_2022,shin_dialogue_2022}, which captures cross-slot shared information by combining slot-specific information descriptions, aiming to achieve knowledge transfer. Furthermore, many studies concentrate on modeling dependencies among slot values to capture information across different domains, while some also utilize large language models and employ methods involving the assembly of domain-slot values for instruction tuning to address the task of dialogue state tracking. These methods allow the model to extend to new domains effectively through autoregressive training, demonstrating better generalization. In this paper, our primary focus lies in zero-shot dialogue state tracking, which involves validating and testing the model in unseen domains after training on visible domains, mainly emphasizing the model's generalization.

\subsection{Parameter-Efficient Transfer-Learning}
With the rapid advancement of large language models, utilizing their capabilities on limited resources has become an urgent problem that needs to be addressed. The current mainstream approach involves parameter-efficient transfer-learning (PETL). This method entails introducing adjustable parameters to a large language model and, during training, freezing the large language model while adjusting only a few parameters, thereby significantly reducing the trainable parameters \cite{houlsby_parameter-efficient_2019,pfeiffer_adapterhub_2020,hu_lora_2021,li_prefix-tuning_2021,lester_power_2021,he_towards_2021,liu_p-tuning_2022,han_ptr_2022,lian_scaling_2022,liu_gpt_2023,ma_parameter-efficient_2023}. In addition to efficiently reducing trainable parameters without compromising model performance, research indicates that parameter-efficient transfer-learning can help avoid overfitting due to the introduction of a few parameters. Moreover, there are many studies applying PETL to dialogue state tracking \cite{madotto_continual_2021,yang_prompt_2022,zhu_continual_2022,wang_divide_2023,aksu_prompter_2023,feng_towards_2023}. Such as \citet{aksu_prompter_2023}, which based on prefix-tuning, introduces a method that dynamically generates prefixes. Besides, \citet{zhu_continual_2022} proposed continual prompt tuning, a parameter-efficient framework that can avoid forgetting and facilitate knowledge transfer between tasks. 
Some research also utilizes large language models with parameter-efficient transfer-learning (PETL) to address dialogue state tracking tasks \cite{feng_towards_2023}. However, since DST is just one crucial part of task-oriented dialogue systems, a weighty component in any part can make it cumbersome and impact overall system responsiveness. This paper proposes a new structure called DualLoRA based on LoRAs for zero-shot dialogue state tracking aiming to fine-tune the model with few parameters and without introducing inference latency.

\section{Dual Low-Rank Adaptation}
This paper proposes DualLoRA, a plug-and-play architecture for dialogue state tracking. DualLoRA consists of a context LoRA and a prompt LoRA. The context LoRA is used to optimize the original dialogue context input, while the prompt LoRA is used to optimize the prompt input.
Both the context LoRA and prompt LoRA are based on LoRA. By employing two distinct LoRAs, the task of zero-shot learning can be divided into two subtasks: tuning the original input and tuning the prompt input. Additionally, to avoid impacting the model's inference latency, the parameters of context LoRA and prompt LoRA can be merged into the pre-trained parameters, which ensures the efficiency of DualLoRA.

\subsection{Context LoRA} 

Adopting a model architecture that minimizes parameter count is crucial for facilitating efficient transfer across domains while maintaining robust performance. Inspired by \citet{hu_lora_2021}, we leverage LoRA with the T5 model to achieve cross-domain transfer in the zero-shot dialogue state task because of LoRA's excellent transferability and its ability to reduce the number of trainable parameters without compromising the model performance.

For context LoRA, we aim to ensure that the model focuses on the dialogue context input in this component. We use a few parameters to tune the model's adaptation to the source domain. By doing so, we can obtain a LoRA that possesses knowledge from the source domain and is easily transferable. 

The context LoRA introduces a bypass structure, enabling the model to fine-tune downstream data without affecting the pre-trained parameters. Context LoRA consists of two matrices: matrix $A$ is initialized through a random Gaussian distribution, and matrix $B$ is initialized with zeros. 
Due to the lower intrinsic dimensionality of pre-trained language models, low-rank decomposition can be applied to constrain the updates of $\Delta W$ for the pre-trained weight matrix $W$, so the updates for $W$ can be formulated as $W + \Delta W = W + BA$, where $A \in {\mathbb{R}^{d \times r}}$, $B \in {\mathbb{R}^{r \times d}}$ and rank $r \ll d$. 
If the input $h \in {\mathbb{R}^d}$, then the output $h^{'}$ can be represented with context LoRA as:

\begin{equation}\label{equation:LoRA}
    {h^{'}} = Wh + \Delta Wh = Wh + BAh
\end{equation}

\begin{figure}[!t]
    \centering
    \includegraphics[width=0.45\textwidth]{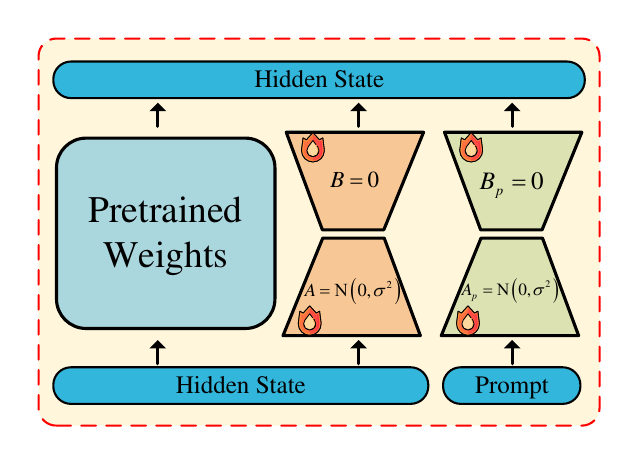}
    \caption{The overall architecture of DualLoRA.}
    \label{fig:architecture}
\end{figure}

\subsection{Prompt LoRA} 

To effectively differentiate and integrate information from the dialogue context and prompts, we propose the incorporation of an additional component to handle the input prompt. Compared with dialogue context, changes in prompts may lead to more significant variations in slot embeddings and hinder the model's adaptability. To handle the variations in the slot embeddings, we use slot prompt \cite{aksu_prompter_2023}, rather than slot embeddings, as the input. 

However, slot prompt introduces noise in the early stage of the training process. To mitigate the noise, we employ the LoRA as the structure for slot prompt tuning, making it a zero-initialized slot prompt. Prompt LoRA consists of two matrices $A_p \in {\mathbb{R}^{d \times r}}$ and $B_p \in {\mathbb{R}^{r \times d}}$. The rank of the two matrices stays the same with the context LoRA. $A_p$ is initialized as Gaussian distribution, while $B_p$ is initialized as zeros. Through zero-vector initialization, when integrating slot prompts, it initially preserves the original knowledge from the pre-trained model. It gradually incorporates prompt-guided signals during the training process, which aids in achieving stable learning during fine-tuning and enhances the model's ability to adhere to prompts in the zero-shot stages.




To train the original input and prompts simultaneously, we combine the context LoRA with the prompt LoRA, resulting in a comprehensive structure that simultaneously accepts prompts $p$ and original inputs $h$. The output $h^{'}$ can be represented as: 

\begin{equation}\label{equation:PromptLoRA_output}
    {h^{'}} = Wh + BAh + {B_p}{A_p}p
\end{equation}

\noindent where $p$ is the slot prompt for prompt LoRA.

Avoiding inference latency is also a noteworthy concern during the inference stage. For the context LoRA, we can directly merge the $B$ and $A$ matrix from the context LoRA into the original pre-trained weights. 
\begin{equation}\label{equation:domain_transfer_inference}
    Wh + BAh = \left( {W + BA} \right)h
\end{equation}
For the prompt LoRA, as its input is different from the context LoRA, it cannot be directly merged into the original pre-trained weights. Therefore, we precompute ${B_p}{A_p}p$ and then merge them into the bias without increasing inference latency so that we can compute $W = W + BA$ and $b = b + {B_p}{A_p}p$. By doing so, we can save the time required during inference.

\subsection{Multi-Head Attention}
After obtaining the output from DualLoRA, we add it to the multi-head attention. In this way, we aim for each head to learn different prompt features, making it easier for prompts to influence the model. We pair DualLoRA with each attention layer in the T5 model, making it easier for prompts to influence the attention mechanism. As the length of the prompt is different from the length of the original input, we adopt the approach of taking the average of the slot prompt embedding and then interacting it with $W_q$ and $W_v$ in the attention mechanism, which is consistent with the weight matrix used in the LoRA \cite{hu_lora_2021}.

\begin{table*}[ht]
\centering
\resizebox{0.9\textwidth}{!}{
\begin{tabular}{@{}cccccccc@{}}
\toprule
\textbf{Model} & \textbf{Pre. Model} & \textbf{Attraction} & \textbf{Hotel} & \textbf{Restaurant} & \textbf{Train} & \textbf{Taxi} & \textbf{Average} \\ \midrule
TRADE          & -                   & 20.06               & 14.20          & 12.59               & 22.39          & 59.21         & 25.69            \\
MA-DST         & -                   & 22.46               & 16.28          & 13.56               & 22.76          & 59.27         & 26.87            \\
SUMBT          & BERT-base           & 22.60               & 19.08          & 16.50               & 22.50          & 59.50         & 28.04            \\
GPT2-DST       & GPT2-base           & 23.67               & 18.54          & 21.05               & 24.34          & 59.10         & 29.34            \\
T5DST          & T5-small            & 31.92               & 20.72          & 20.09               & 28.83          & 64.12         & 33.14            \\
SlotDM-DST     & T5-small            & 33.92               & 18.85          & 20.75               & 36.96          & 66.25         & 35.35            \\
T5DST*          & PPTOD-small         & 35.5                & 20.0           & 25.3                & 35.3           & 65.6          & 36.4             \\
Prompter       & PPTOD-small         & 35.8       & 19.2           & 26.0                & 39.0           & 66.3          & 37.2            \\
DCC            & T5-small            & 35.8                & \textbf{24.8}  & 22.9                & 40.2           & 65.9          & 37.9             \\
DualLoRA (Ours)   & PPTOD-small         & \textbf{37.1}                & 18.9           & \textbf{27.9}       & \textbf{42.4}  & \textbf{67.2} & \textbf{38.7}   \\ \bottomrule
\end{tabular}
}
\caption{Zero-shot joint-goal accuracy(\%) on MultiWOZ. All results of baselines are reported from original papers. The best results in each column are bold. Pre. Model means the pretrained model. The Model with * is excerpted from Prompter \cite{aksu_prompter_2023}.}
\label{table: MultiWOZ}
\end{table*}

\begin{table*}[ht]
\centering
\resizebox{0.95\textwidth}{!}{
\begin{tabular}{@{}ccccccccc@{}}
\toprule
\textbf{Model} & \textbf{Buses}     & \textbf{Events} & \textbf{Flights}       & \textbf{Media}     & \textbf{Messaging} & \textbf{Music}     & \textbf{Payment}   & \textbf{Trains}    \\ \midrule
SGD-baseline   & 9.7/50.9           & 23.5/57.9       & 23.9/65.9         & 18.0/30.8          & 10.2/20.0          & 15.5/39.9          & 11.5/34.8          & 13.6/63.5          \\
Seq2seq-DU     & 16.8/N             & 31.9/N          & 15.9/N            & 23.1/N             & 4.9/N              & 12.3/N             & 7.2/N              & 16.8/N             \\
Transfer-QA    & 15.9/63.6          & 15.6/56.8       & 3.59/42.9         & 30.2/67.5          & 13.3/37.9          & 8.9/62.4           & \textbf{24.7/60.7} & 17.4/64.9          \\
SlotDM-DST     & 43.9/86.3          & -               & -                  & -                  & 36.6/61.4          & -                  & 16.5/62.0          & 46.7/86.9          \\
T5DST*        & 46.8/N             & 48.8/N          & -                  & 55.5/N             & 59.2/N             & -                  & 23.3/N             & 53.0/N             \\
PROMPTER       & 48.4/N             & \textbf{51.5/N} & -                  & 65.3/N             & 59.2/N             & -                  & 21.9/N             & 50.8/N             \\
DCC            & -                  & -               & -                  & -                  & 28.8/N             & -                  & 19.4/N             & 42.3/N             \\
DualLoRA (Ours)           & \textbf{50.9/88.8} & 46.5/82.8       & \textbf{28.4/76.9}   & \textbf{69.7/88.7} & \textbf{65.1/85.5} & \textbf{32.5/72.4} & 21.2/70.2          & \textbf{52.9/89.3} \\ \bottomrule
\end{tabular}
}
\caption{Zero-shot results on SGD dataset. All results are reported in JGA(\%)/AGA(\%). N represents the results are not reported in the original paper. The Model with * is excerpted from Prompter  \cite{aksu_prompter_2023}.}
\label{table: SGD}
\end{table*}

\begin{table}[!t]
\centering
\begin{adjustbox}{width=0.45\textwidth,scale=1}
\begin{tabular}{@{}lccccc@{}}
\toprule
\textbf{Model}       & \textbf{Attraction} & \textbf{Hotel} & \textbf{Train} & \textbf{Taxi} & \textbf{Restaurant} \\ \midrule
T5DST               & 35.5                & \textbf{20.0}  & 35.3           & 65.6          & 25.3                \\

+ ContextLoRA & 36.0       & 17.0           & 41.2           & 66.9          & 26.8                \\
+ PromptLoRA                & \textbf{37.1}                & 18.9           & \textbf{42.4}  & \textbf{67.2} & \textbf{27.9}       \\ \bottomrule
\end{tabular}
\end{adjustbox}
\caption{Ablation results on the test set of MultiWOZ.}
\label{table: ablation}
\end{table}

\section{Experiments}
\subsection{Dataset and Metrics}
We conduct our experiments on the MultiWOZ dataset and Schema-Guided-Dialogue (SGD) dataset \cite{budzianowski_multiwoz_2018,rastogi_towards_2020,eric_multiwoz_2020}, which are well-known benchmarks in task-oriented dialogue systems. MultiWOZ is an extensively annotated dataset comprising 10,000 human-to-human written conversations covering diverse domains and topics, making it a widely used benchmark dataset for evaluating task-oriented dialogue systems. The SGD dataset, like MultiWOZ, is extensively annotated with turn-level information and descriptions, which comprises over 20,000 annotated dialogues between a human and a virtual assistant, covering more than 20 domains. Additionally, the SGD dataset includes previously unseen domains in the test set, designed to measure the zero-shot performance of dialogue systems.

In zero-shot cross-domain dialogue state tracking, the model is initially trained on visible domains and tested on unseen domains. Building on prior work, we use Joint Goal Accuracy (JGA) and Average Goal Accuracy (AGA) to evaluate both the baseline and our model. JGA refers to the percentage of turns for which all the slots in the same domain are identified correctly, which is the primary metric for dialogue state tracking evaluation and is used to evaluate the model performance. AGA refers to the average accuracy of the active slots in each turn. The evaluation metrics used are consistent with Prompter \cite{aksu_prompter_2023}.

\subsection{Baselines}
We compare the proposed method with TRADE \cite{wu_transferable_2019}, MA-DST \cite{kumar_ma-dst_2020}, SUMBT \cite{lee_sumbt_2019}, SGD-baseline \cite{rastogi_scaling_2019}, Seq2Seq-DU \cite{feng_sequence--sequence_2021}, GPT2-DST \cite{li_zero-shot_2021}, TransferQA \cite{lin_zero-shot_2021}, T5DST \cite{lin_leveraging_2021}, SlotDM-DST \cite{wang_slot_2022}, DCC \cite{wang_divide_2023}  and Prompter \cite{aksu_prompter_2023}. More details about baselines are presented in Appendix \ref{sec: baseline}.

\subsection{Implementation Details}
To facilitate a better comparison, the adopted settings are consistent with those of Prompter \cite{aksu_prompter_2023}. We utilize T5 architecture as our backbone. As mentioned in Prompter, PPTOD \cite{su_multi-task_2022} is more suitable for prompt tuning tasks because of the nature of its pre-training task. Therefore we choose PPTOD-small as our checkpoint. For more detail, the batch size is set to 8 with gradient accumulation per 8 steps; the optimizer is set to AdamW \cite{loshchilov_decoupled_2018}, and the learning rate is set to 1e-4. All experiments were conducted on the NVIDIA GeForce RTX 3090 and 4090 GPUs.

\subsection{Main Results}
As shown in Table \ref{table: MultiWOZ}, DualLoRA performs best except hotel domain compared to other baselines. That is because in the hotel domain, the presence of "none" value slots and unique value slots make it challenging for the model to utilize prompts for zero-shot dialogue state tracking effectively. In addition, the slots in the hotel domain are unique and cannot be effectively utilized using domain prompts during training. Moreover, we observe that the model can learn how to effectively utilize prompts between attraction and restaurant domains due to the close relationship between slot values in these two domains.  Consequently, they influence each other, leading to better performance of the model in attraction and restaurant domains. Similarly, the train and taxi domains follow a similar pattern.

As shown in Table \ref{table: SGD}, we present the performance of different models on various domains in the SGD task, which can be observed that DualLoRA exhibits the best performance in most domains. These domains are those for which the service in the test set is not present in the training set. The improvements of DualLoRA are attributed to effectively training on prompts in visible domains and leveraging prompt utilization in unseen domains.  Due to the pretraining of the alarm domain in PPTOD, we do not report the scores of our model on the alarm domain in this experiment.

\begin{figure}[!t]
    \centering
    \includegraphics[width=0.35\textwidth]{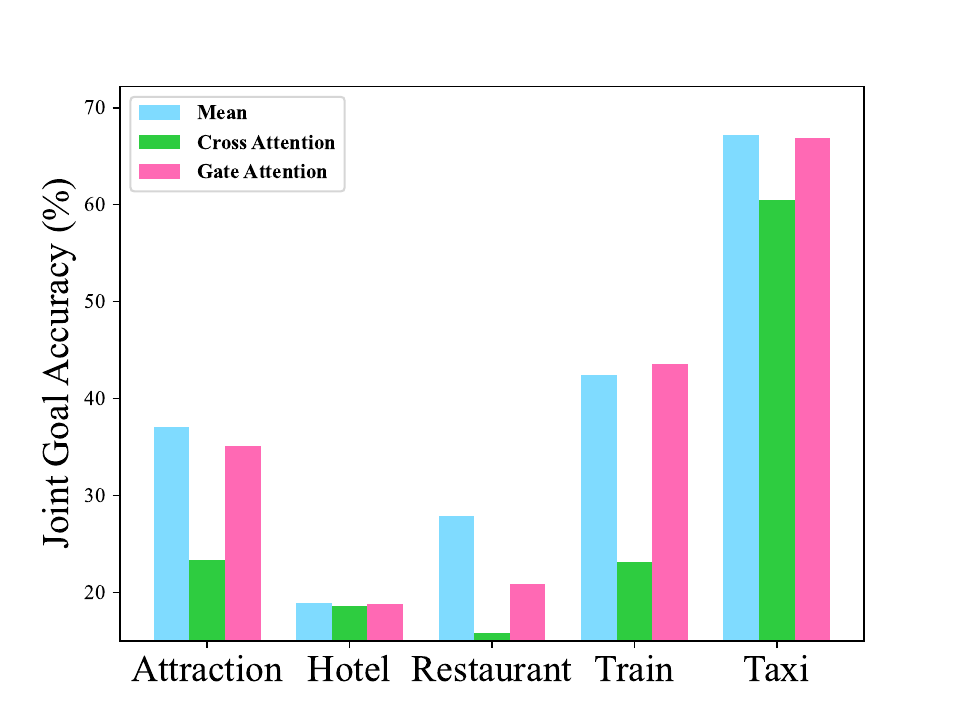}
    \caption{Performance of DualLoRA with different fusion methods on the MultiWOZ dataset.}
    \label{fig: fusion method}
\end{figure}

\begin{figure}[!t]
    \centering
    \includegraphics[width=0.35\textwidth]{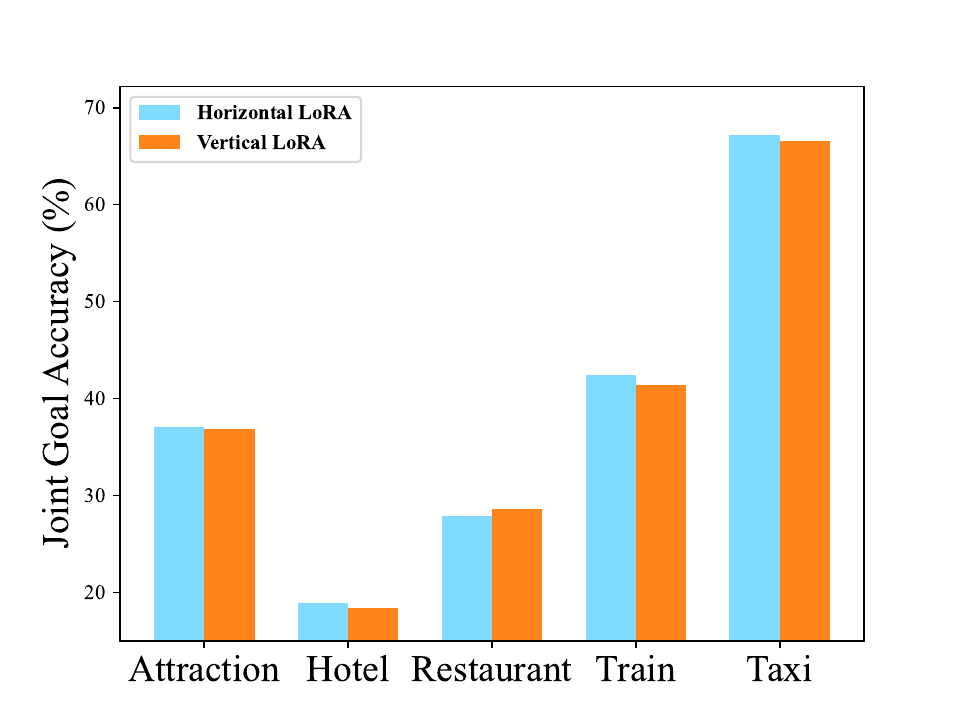}
    \caption{Performance of DualLoRA at different positions on the MultiWOZ dataset.}
    \label{fig: horizontal and veritcal}
\end{figure}

\begin{figure}[!t]
    \centering
    \includegraphics[width=0.35\textwidth]{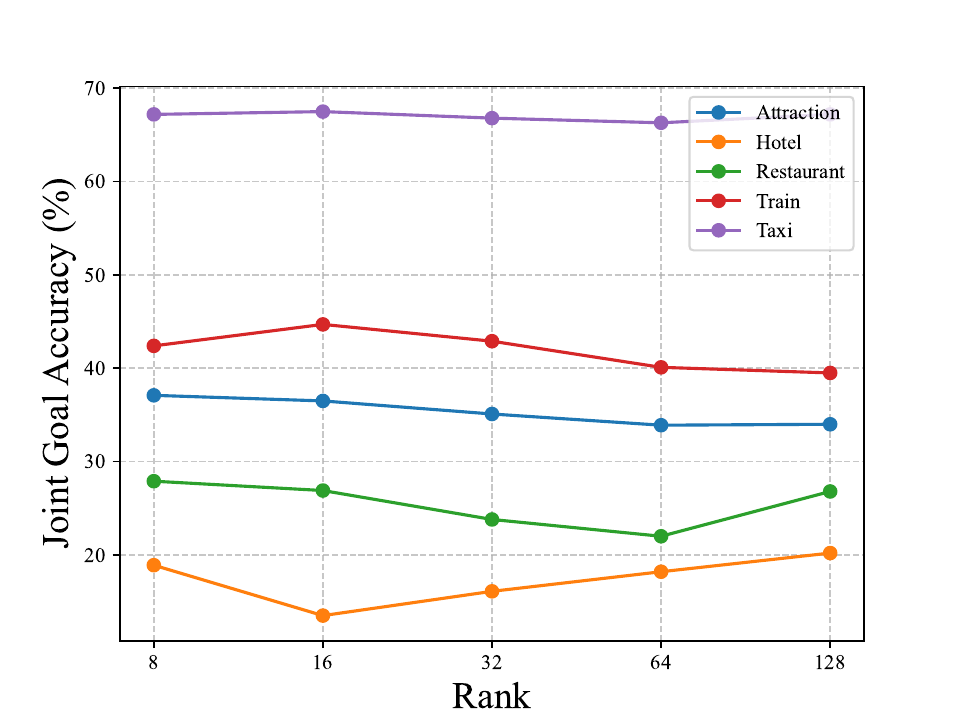}
    \caption{Performance of DualLoRA with different rank on the MultiWOZ dataset.}
    \label{fig: rank}
\end{figure}

\subsection{Ablation Study}
To investigate the effectiveness of different components in our model, we conduct an ablation experiment on context LoRA and prompt LoRA. Table \ref{table: ablation} demonstrates the effectiveness of the two components in our method. After adding the context LoRA, the model shows improvement across multiple domains, indicating that this structure plays a crucial role in the model's generalization. After adding prompt LoRA, the model also exhibits improvement across multiple domains, which clearly illustrates the contribution of the proposed structure.

\begin{table}[!t]
\centering
\resizebox{0.45\textwidth}{!}{
\begin{tabular}{@{}cccccc@{}}
\toprule
      \textbf{Model} & \textbf{Attraction} & \textbf{Hotel} & \textbf{Taxi} & \textbf{Train} & \textbf{Restaurant} \\ \midrule
      w/ MLP         & 21.3                & 10.5           & 57.0          & 20.7           & 14.1                \\
      w/ Adapter     & 31.7                & 18.4           & 65.0          & 39.0           & 20.4                \\
      w/ Transformer & 22.2                & 11.6           & 64.0          & 25.8           & 14.9                \\
      w/ LoRA        & \textbf{37.1}       & \textbf{18.9}  & \textbf{67.2}  & \textbf{42.4}  & \textbf{27.9}       \\ \bottomrule
    \end{tabular}
}
\caption{Different bypass structures employed in DualLoRA.}
\label{table: bypass}
\end{table}

\subsection{Analysis}
Moreover, we investigate the impact of different types of the bypass structure. We tested four different structures, including MLP, Adapter, Transformer, and LoRA. Results in Table \ref{table: bypass} show that LoRA excels in handling prompt information, and in terms of parameter usage, LoRA has the smallest number of parameters.

\begin{figure}[!t]
    \centering
    \includegraphics[width=0.35\textwidth]{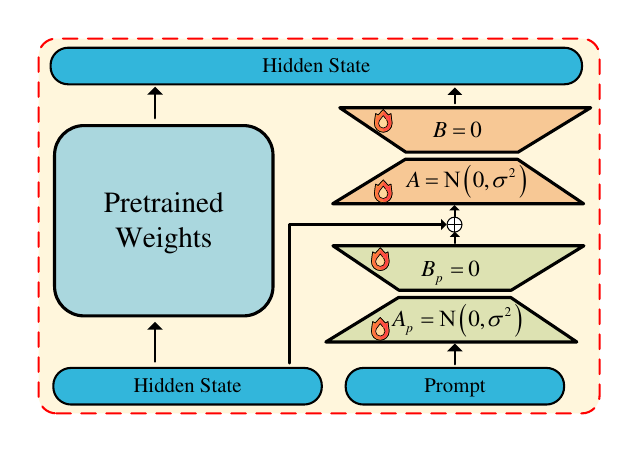}
    \caption{Vertical Combination of PromptLoRA and ContextLoRA.}
    \label{fig: vertical_lora}
\end{figure}

To explore the optimal fusion position of DualLoRA, we also tested two fusion positions for context LoRA and prompt LoRA: vertical combination and horizontal combination. The horizontal combination is the one we proposed in Figure \ref{fig:architecture}, while the vertical combination is shown in Figure \ref{fig: vertical_lora}. As indicated in Figure \ref{fig: horizontal and veritcal}, horizontal combination yields better results than vertical combination. We attribute the inferior performance of the vertical combination to the fact that a single context LoRA may not effectively optimize both types of information simultaneously.

We employed various fusion methods to explore the optimal fusion approach, such as mean addition, cross-attention interaction, and gate-attention interaction. As shown in Figure \ref{fig: fusion method}, in most cases, gate-attention and mean addition are superior to cross-attention interaction, which may be attributed to the initial dialogue context having a weak influence on prompt interaction, making it challenging for the model to converge effectively. In addition, both cross-attention and gate-attention introduce many additional parameters, which may not be favorable for the model in domain transfer scenarios. Therefore, we adopt mean addition as the fusion method for our different inputs.

\begin{figure}[!t]
    \centering
    \includegraphics[width=0.35\textwidth]{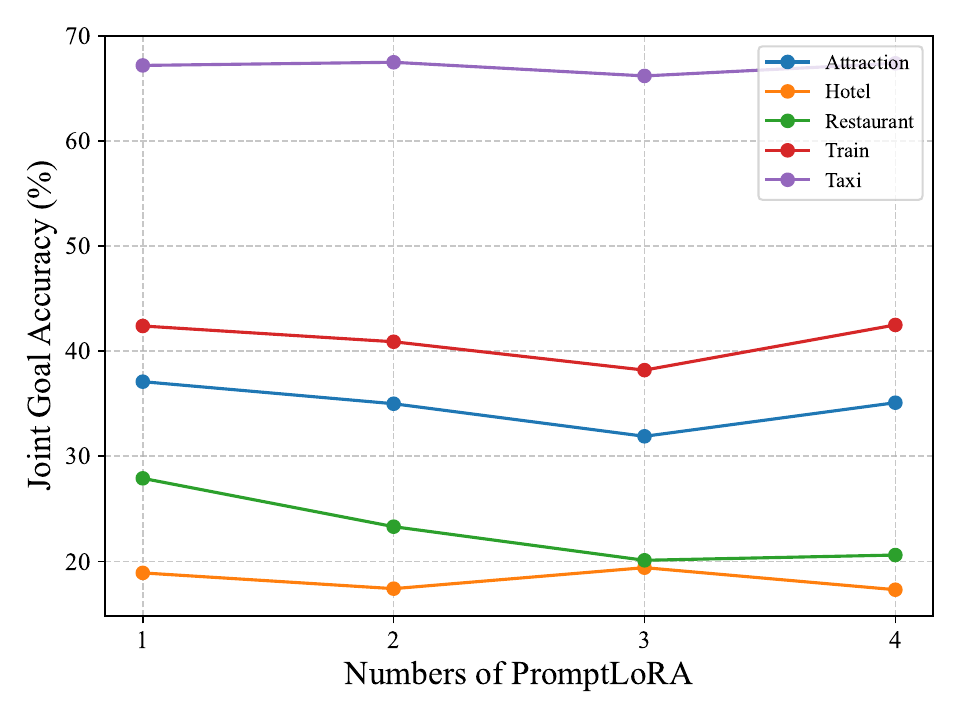}
    \caption{Performance of different numbers of prompt LoRAs on the MultiWOZ dataset.}
    \label{fig: numofplora}
\end{figure} 

We also conducted experiments to explore the optimal matrix rank of DualLoRA since it determines the matrix rank, including 8, 16, 32 and 64. As shown in Figure \ref{fig: rank}, the performance of the model does not consistently improve with the increase of rank, which is in line with the observation by \cite{li_measuring_2018,aghajanyan_intrinsic_2021} that the learned over-parametrized models reside on a low intrinsic dimension. Therefore, we adopt a rank of 8 for DualLoRA in all experiments.

In order to investigate the impact of the number of DualLoRA on model performance, we conducted experiments. The purpose was to explore whether the utilization of prompts becomes more effective with an increasing number of prompt LoRA. The results are shown in the Figure \ref{fig: numofplora}. Our analysis suggests that the limited size of the model parameters may be the reason behind the lack of significant performance improvement with an increasing number of prompt LoRA. Additionally, the introduction of more prompt LoRA leads to an increase in the overall parameter count. Therefore, in this study, we opted to utilize only one prompt LoRA to enhance the model's utilization of prompts.


\begin{figure}[!t]
    \centering
    \label{fig:enter-label}
\begin{subfigure}{0.21\textwidth}
        \includegraphics[width=\linewidth]{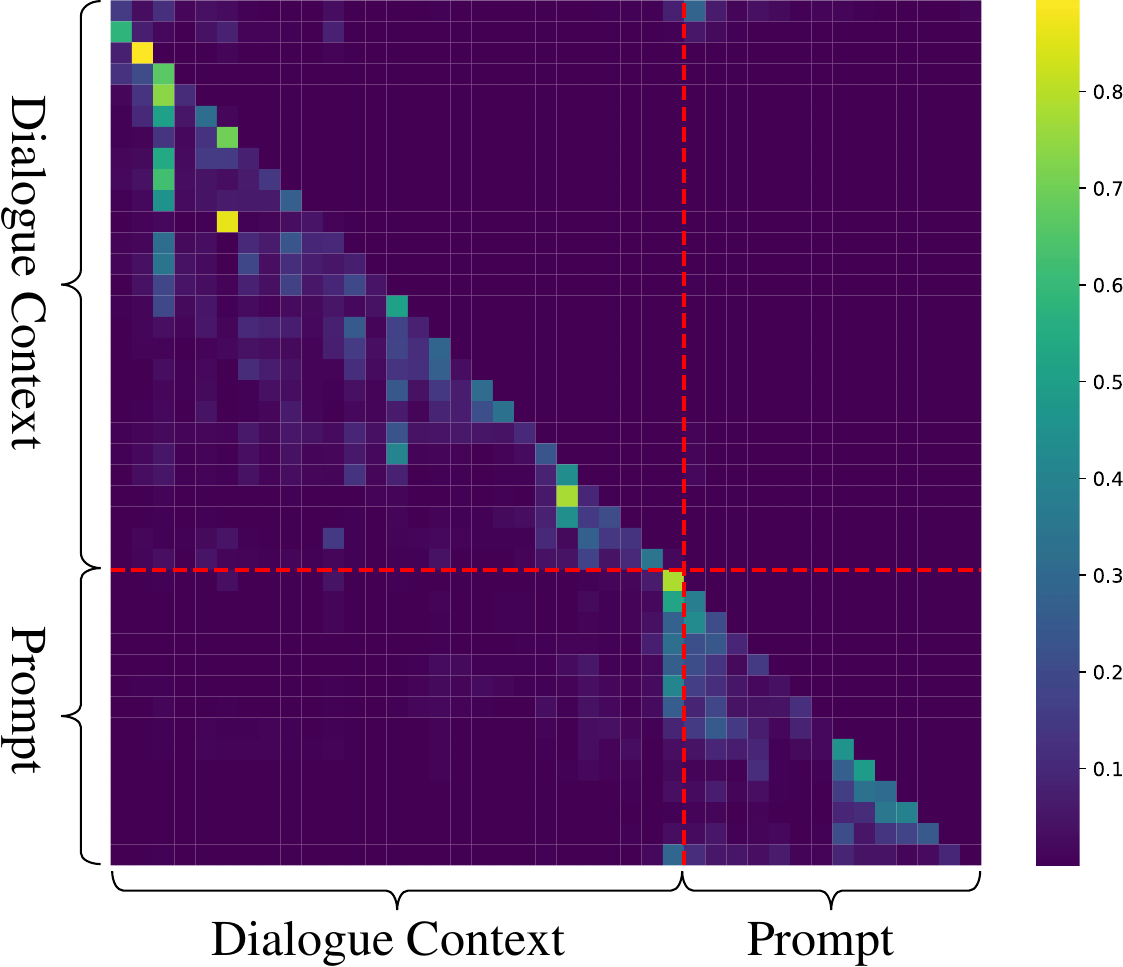}
        \caption{T5DST}
\end{subfigure}
\begin{subfigure}{0.215\textwidth}
        \includegraphics[width=\linewidth]{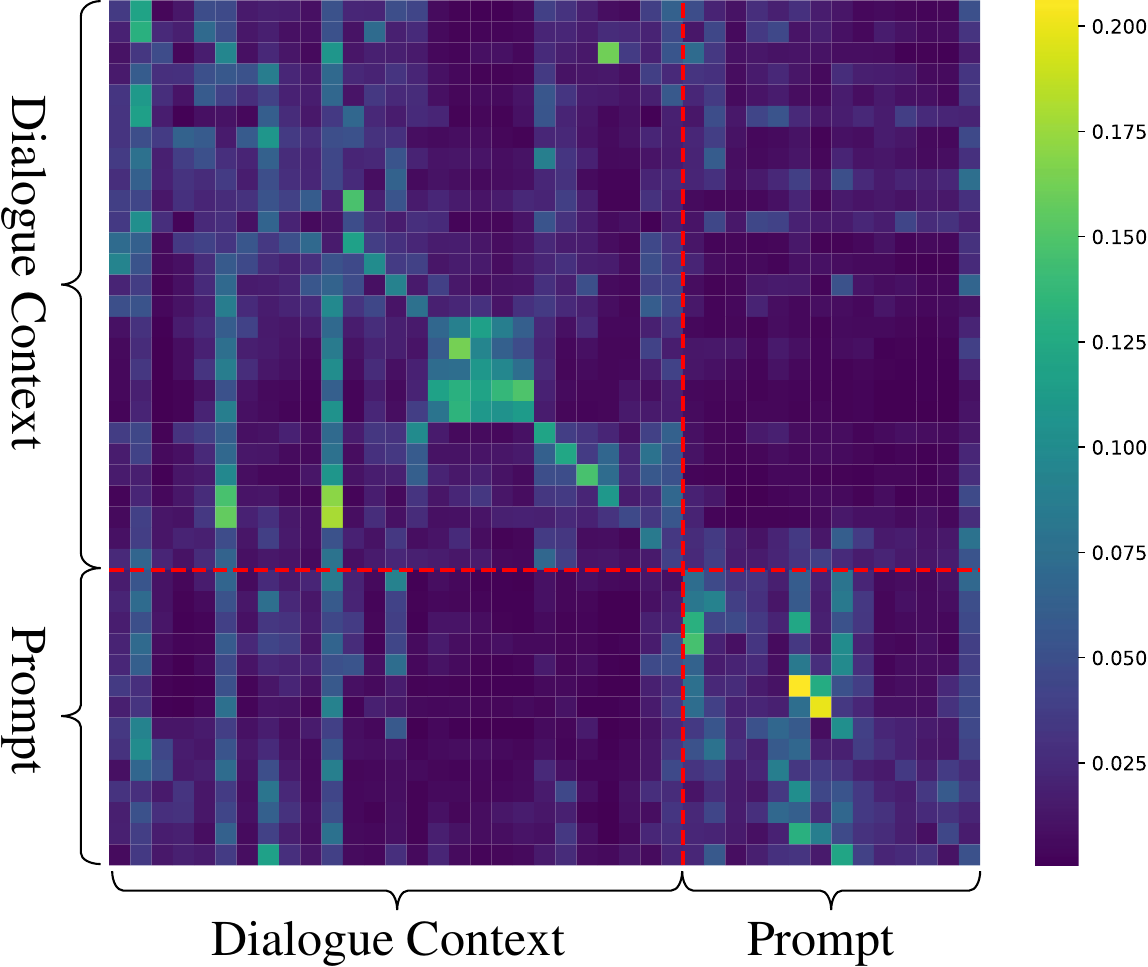}
        \caption{DualLoRA}
\end{subfigure}
\caption{Example attention maps of the last encoder layer in T5DST and DualLoRA.}
\label{fig: after_attn}
\end{figure}

To validate the effectiveness of our method, we visualized the attention in the last layer of the model's encoder to explore the model's focus on the dialogue context and prompts. As shown in Figure \ref{fig: after_attn}, we can observe that, the dialogue context pays more attention to the prompts compared to Figure \ref{fig: attn_map}. In addition, the attention scores in Figure \ref{fig: attn_map} after softmax are more extreme, ranging from 0 to 1, and are mainly distributed along the main diagonal, indicating a focus on itself. In contrast, Figure \ref{fig: after_attn} demonstrates a more diverse attention distribution, with effective attention on prompts. This observation further supports the effectiveness of our proposed method.

\subsection{Case Study}
\begin{figure}[!t]
    \centering
    \includegraphics[width=0.45\textwidth]{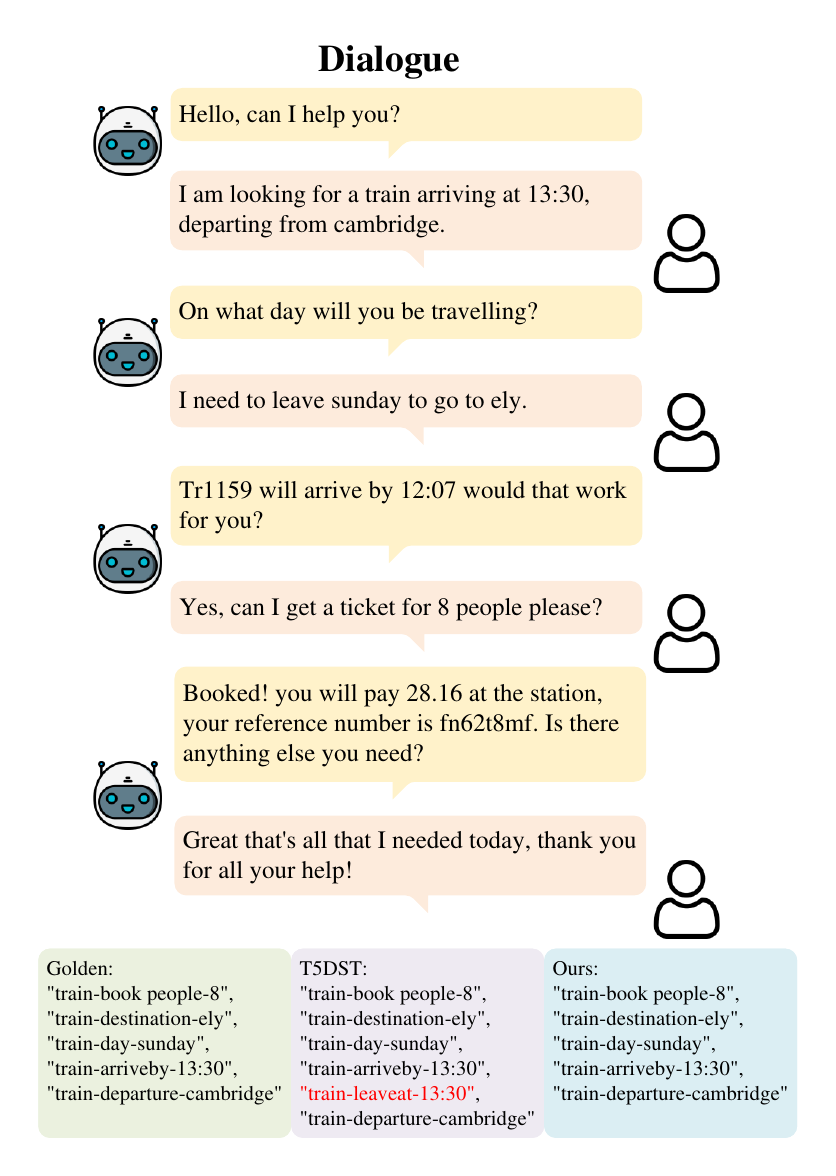}
    \caption{Zero-shot dialogue state tracking based on different models in MultiWOZ.}
    \label{fig: case study}
\end{figure}

We selected a dialogue example from the MultiWOZ dataset to demonstrate the performance of our model compared to T5DST in zero-shot dialogue state tracking, as shown in Figure \ref{fig: case study}. The results show that both T5DST and our model predict most zero-shot dialogue states in the train domain, such as "train-book people-8" and "train-destination-ely." The rationale is that the models have been trained in similar domains, such as the taxi domain, allowing them to provide better answers. Moreover, the results show that T5DST, during zero-shot inference in the train domain, incorrectly equates the "leaveat" slot value with the "arriveby" slot value and predicts it as "13:30", which is because T5DST fails to effectively utilize prompt information, making it unable to differentiate between different slot values for "leaveat" and "arriveby" accurately.

\section{Conclusion}
In summary, this paper introduces DualLoRA based on a parameter-efficient transfer-learning method for zero-shot dialogue state tracking. DualLoRA consists of two parts, namely context LoRA and prompt LoRA. DualLoRA can simultaneously accept two inputs: normal input and prompt input. Through experiments, we demonstrate that this method improves the Joint Goal Accuracy (JGA) metric for MultiWOZ and SGD datasets without introducing inference latency. Additionally, we conduct a series of experiments to show the effectiveness of deploying DualLoRA and how to fuse two different types of information best. Our analysis indicates that the effectiveness of DualLoRA stems from the model's ability to extract and utilize prompt information efficiently.

Future work will focus on applying DualLoRA to various domains, expanding the method to multimodal zero-shot dialogue state tracking, or researching better fusion methods for more effective integration of prompts and dialogue context.

\section{Limitations}
DualLoRA has the following limitations currently. Addressing these limitations in future work will contribute to the broader applicability of DualLoRA.

\begin{itemize}
    \item \noindent Drawbacks of precomuptation. The reason we can use precomputation to reduce inference latency is that the slot prompts in the current target domain are limited. However, as the number of domains to be transferred increases, the computation introduced by precomputation will also grow. Therefore, the inference latency will also increase.

    \item \noindent Unable to be applied to target domains without prompts. Due to DualLoRA relying on enhancing the model's utilization of prompts, if there are no prompts in the target domain, DualLoRA cannot leverage prompts to improve the model's performance.
    
\end{itemize}

\section*{Acknowledgements}
This work was supported by the National Natural Science Foundation of China (NSFC) under Grant Nos. 61966038 and 62266051, the Yunnan University Talent Research Startup Support Project under grant No. CY22623101 and No. CZ22623101, the Postgraduate Research and Innovation Foundation of Yunnan University under Grant No. KC-23234274 and the Exam-Exempted Postgraduate Research and Innovation Foundation of Yunnan University under Grant No. TM-23236972. The authors would like to thank the anonymous reviewers for their constructive comments. The authors would like to thank Prompter \cite{aksu_prompter_2023} for providing the publicly available code.


\bibliographystyle{acl_natbib}
\bibliography{custom}

\newpage

\appendix
\section{Dataset Statistic}

The statistical data for the MultiWOZ dataset is presented in Table \ref{table: MultiWOZ statistic}. Due to the limited data in the police and hospital domains within MultiWOZ, we conducted experiments using only five domains from the dataset. The statistical data for the SGD dataset is shown in Table \ref{tab: SGD statistic}.

\begin{table}[h]
\centering
\begin{tabular}{@{}cccc@{}}
\toprule
\textbf{Domain} & Train & Dev  & Test \\ \midrule
Attraction      & 2717  & 401  & 416  \\
Hotel           & 3381  & 416  & 394  \\
Restaurant      & 3813  & 438  & 207  \\
Taxi            & 1654  & 207  & 195  \\
Train           & 3103  & 484  & 494  \\
Total           & 8438  & 1000 & 1000 \\ \bottomrule
\end{tabular}
\caption{The dataset statistic of MultiWOZ.}
\label{table: MultiWOZ statistic}
\end{table}

\begin{table}[h]
\centering
\resizebox{0.45\textwidth}{!}{
\begin{tabular}{@{}cc|cc@{}}
\toprule
\textbf{Domain}    & Total Dialogues & \textbf{Domain}      & Total Dialogues \\ \midrule
Alaram    & 324             & Movies      & 2339            \\
Banks     & 1021            & Music       & 1833            \\
Buses     & 3135            & Payment     & 222             \\
Calendar  & 1602            & RentalCars  & 2510            \\
Events    & 4519            & Restaurants & 3218            \\
Flights   & 3644            & RideSharing & 2223            \\
Homes     & 1273            & Services    & 2956            \\
Hotels    & 4992            & Trains      & 350             \\
Media     & 1656            & Travel      & 2808            \\
Messaging & 298             & Weather     & 1783            \\ \bottomrule
\end{tabular}
}
\caption{The dataset statistic of SGD.}
\label{tab: SGD statistic}
\end{table}

\section{Baseline Models}\label{sec: baseline}

In this section, we provide a detailed overview of each baseline, as outlined below.

\begin{itemize}
    \item \noindent {\bf TRADE}. TRADE \cite{wu_transferable_2019} aims to enhance dialogue state generation by incorporating a copy mechanism and enabling knowledge transfer for previously unseen dialogue states during model training.
    
    \item \noindent {\bf MA-DST}. MA-DST \cite{kumar_ma-dst_2020} utilizes cross-attention to model relationships between context and slots across different semantic levels and employs self-attention to resolve cross-domain coreferences based on the RNN layer. 
    
    \item \noindent {\bf SUMBT}. SUMBT \cite{lee_sumbt_2019} learns relationships between domain-slot-types and slot-values through attention mechanisms based on contextual semantic vectors, predicting slot-value labels in a non-parametric manner.

    \item \noindent {\bf SGD-baseline}. SGD-baseline \cite{rastogi_towards_2020} encodes these elements into embedded representations, accommodating dynamic sets of schema elements by conditioning on corresponding schema embeddings based on the BERT model to predict dialogue state.

    \item \noindent {\bf Seq2Seq-DU}. Seq2Seq-DU \cite{feng_sequence--sequence_2021} treats DST as a sequence-to-sequence problem and employs two BERT-based encoders to encode dialogue utterances and schema descriptions independently. After that, Seq2Seq-DU uses a decoder to generate a pointer representation for the current dialogue state.

    \item \noindent {\bf GPT2-DST}. GPT2-DST \cite{li_zero-shot_2021} utilizes a generative question-answering model pre-trained on English sentences, enabling natural language queries for unseen constraints and slots in multi-domain task-oriented dialogues.

    \item \noindent {\bf TransferQA}. TransferQA \cite{lin_zero-shot_2021} is a transferable generative QA model that seamlessly integrates extractive and multiple-choice QA using a text-to-text transformer framework, which effectively tracks both categorical and non-categorical slots in DST and introduces innovative methods to construct unanswerable questions.
    
    \item \noindent {\bf T5DST}. T5DST \cite{lin_leveraging_2021} encodes dialogue context and slots using a pre-trained self-attentive encoder, generating slot values in an auto-regressive manner. Additionally, the model incorporates slot-type informed descriptions to capture shared information across slots, facilitating cross-domain knowledge transfer. 

    \item \noindent {\bf SlotDM-DST}. SlotDM-DST \cite{wang_slot_2022} employs slot prompts combination, slot values demonstration, and slot constraint objects to model slot-slot, slot-value, and slot-context dependencies, respectively. Each slot prompt includes a slot-specific and a slot-shared prompt to capture shared knowledge across domains.

    \item \noindent {\bf DCC}. Divide, Conquer and Combine (DCC) \cite{wang_divide_2023} employs a "divide, conquer, and combine" strategy, explicitly disentangling semantics through a mixture-of-experts mechanism. Seen data is divided into semantically independent subsets, corresponding experts are trained, and unseen samples are inferred using the mixture-of-experts with ensemble inference.
    
    \item \noindent {\bf Prompter}. Prompter \cite{aksu_prompter_2023} addresses this by utilizing descriptions of target domain slots to generate dynamic prefixes. These prefixes are then concatenated to the key and values at each layer's self-attention mechanism, enabling prefix-tuning in zero-shot scenarios. 

\end{itemize}

\section{Best Applied Weight Matrices}
To explore where we should apply DualLoRA in T5's weight matrices, we conducted experiments applying DualLoRA to the attention layers, such as query, key, value, and output.  As shown in the table \ref{table: qkvo}, in most domains, achieving the best performance is possible by solely utilizing the attention layers for queries (q) and values (v), consistent with the findings in the LoRA paper \cite{hu_lora_2021}. Moreover, using only the q and v attention layers reduces the model's parameters. Therefore, in our experiments, we opted for applying DualLoRA only to the q and v attention layers.

\begin{table}[h]
\centering
\resizebox{0.45\textwidth}{!}{
\begin{tabular}{@{}cccccc@{}}
\toprule
\textbf{Weight Matrix} & Attraction    & Hotel         & Restaurant    & Train         & Taxi          \\ \midrule
qv                     & \textbf{37.1} & 18.9          & \textbf{27.9} & \textbf{42.4} & \textbf{67.2} \\
qkv                    & 33.9          & 17.4          & 23.4          & 40.5          & 67.2          \\
qkvo                   & 32.2          & \textbf{19.9} & 20.9          & 39.5          & 66.5          \\ \bottomrule
\end{tabular}
}
\caption{Different types of attention weights employed in DualLoRA.}
\label{table: qkvo}
\end{table}

\section{Slot Performance on MultiWOZ}
\begin{figure}[t!]
    \centering
    \includegraphics[width=0.49\textwidth]{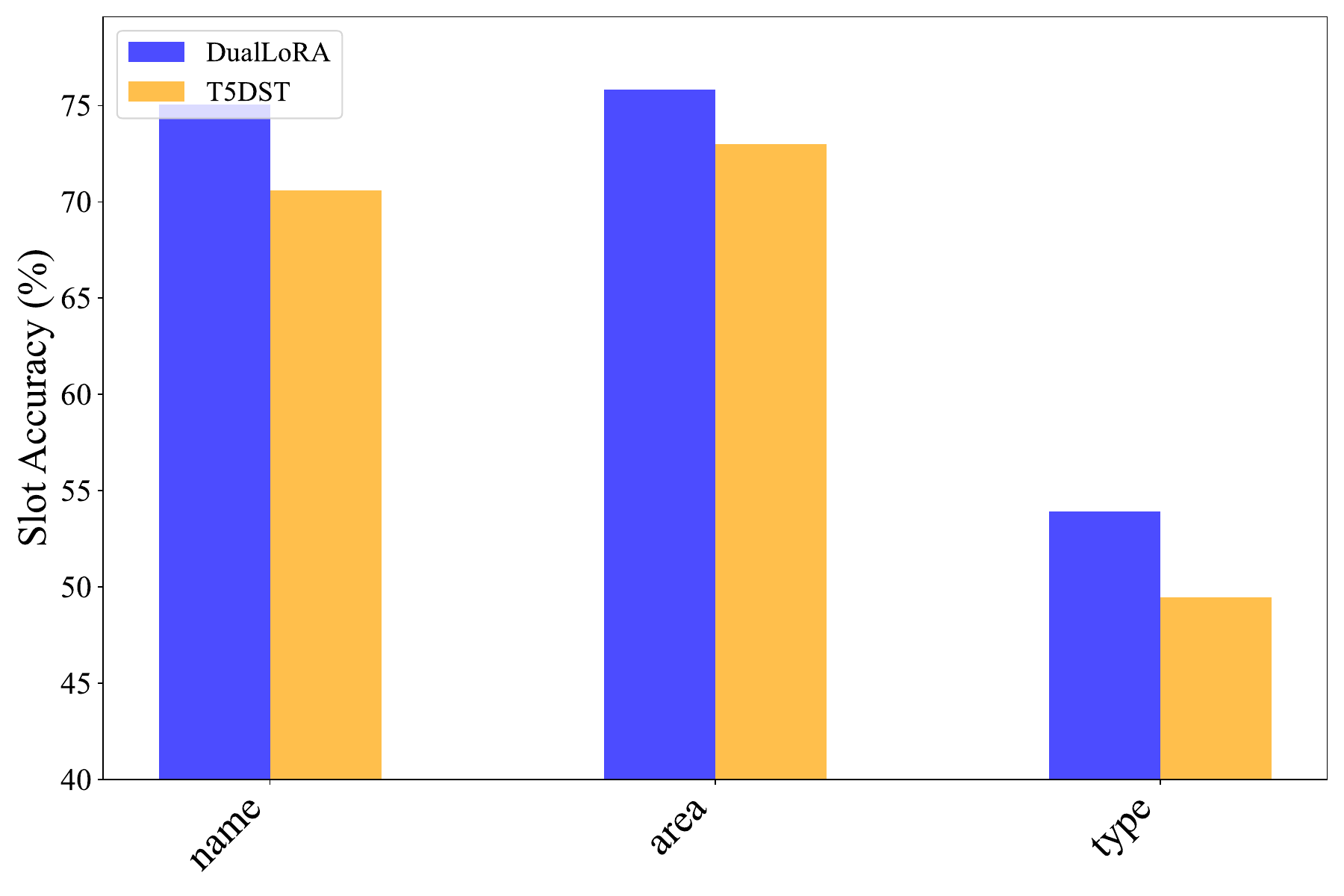}
    \caption{Slot Accuracy in attraction domain on MultiWOZ dataset.}
    \label{fig: attraction_acc}
\end{figure}

\begin{figure}[t!]
    \centering
    \includegraphics[width=0.49\textwidth]{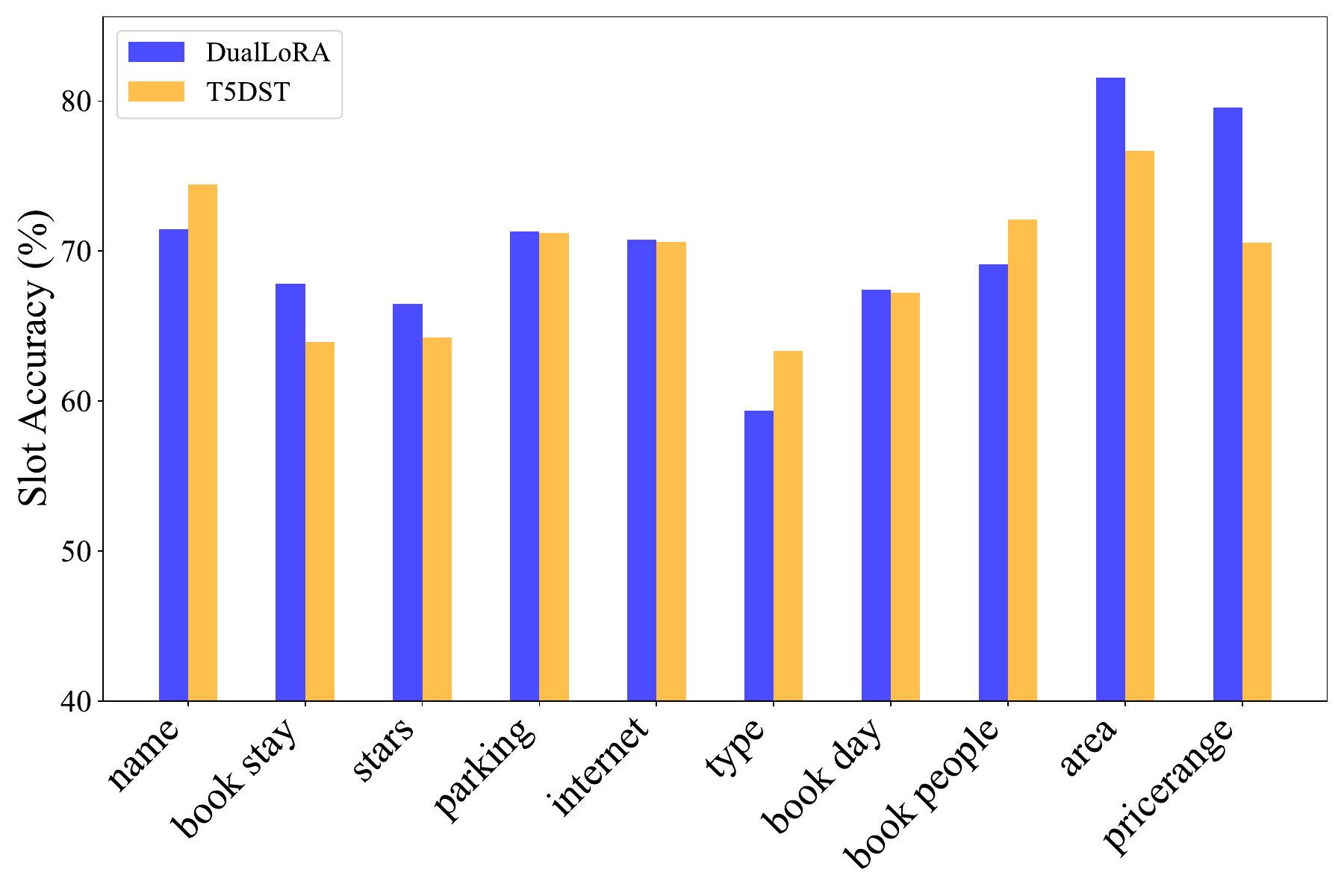}
    \caption{Slot Accuracy in hotel domain on MultiWOZ dataset.}
    \label{fig: hotel_acc}
\end{figure}

\begin{figure}[t!]
    \centering
    \includegraphics[width=0.49\textwidth]{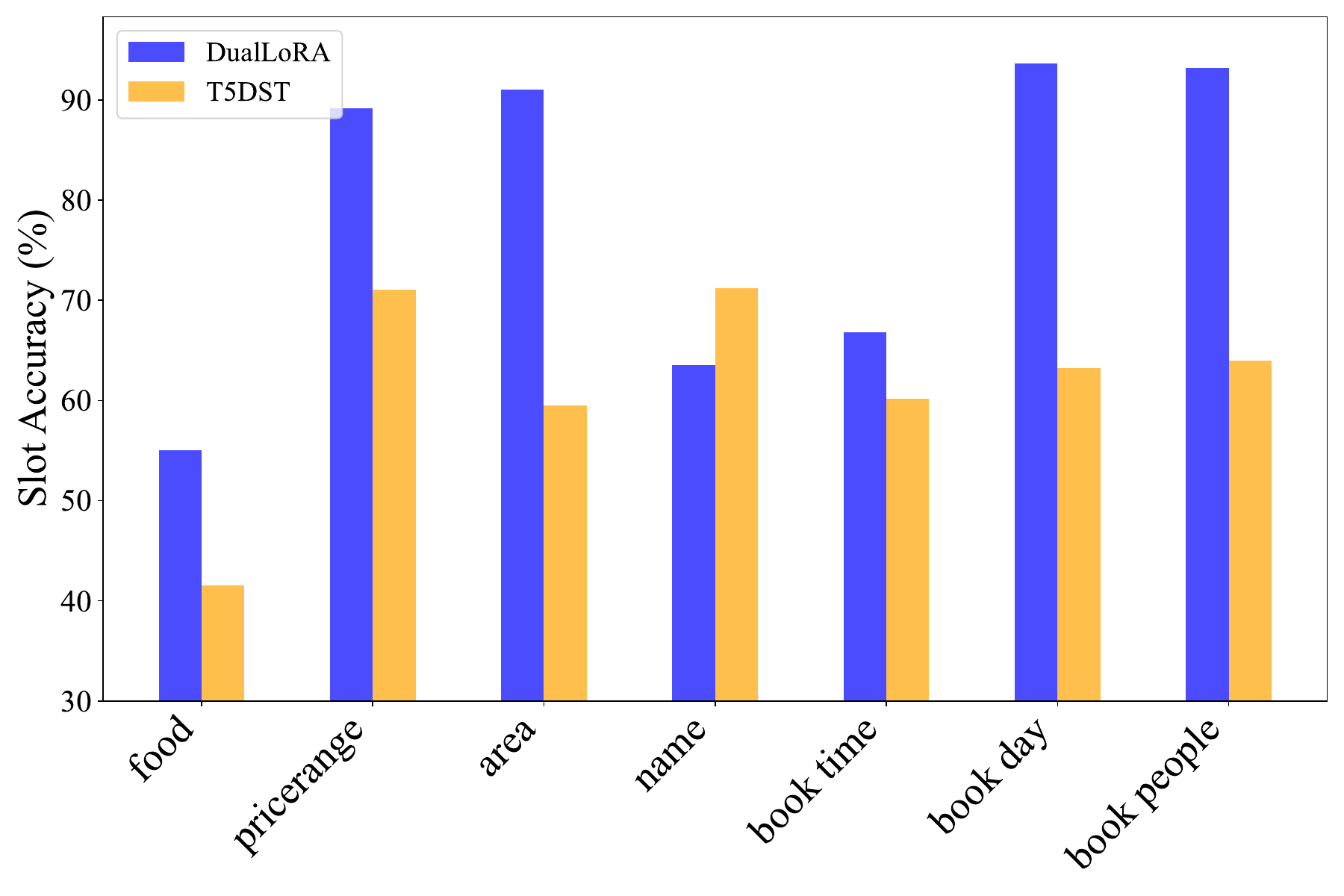}
    \caption{Slot Accuracy in restaurant domain on MultiWOZ dataset.}
    \label{fig: restaurant_acc}
\end{figure}

\begin{figure}[t!]
    \centering
    \includegraphics[width=0.49\textwidth]{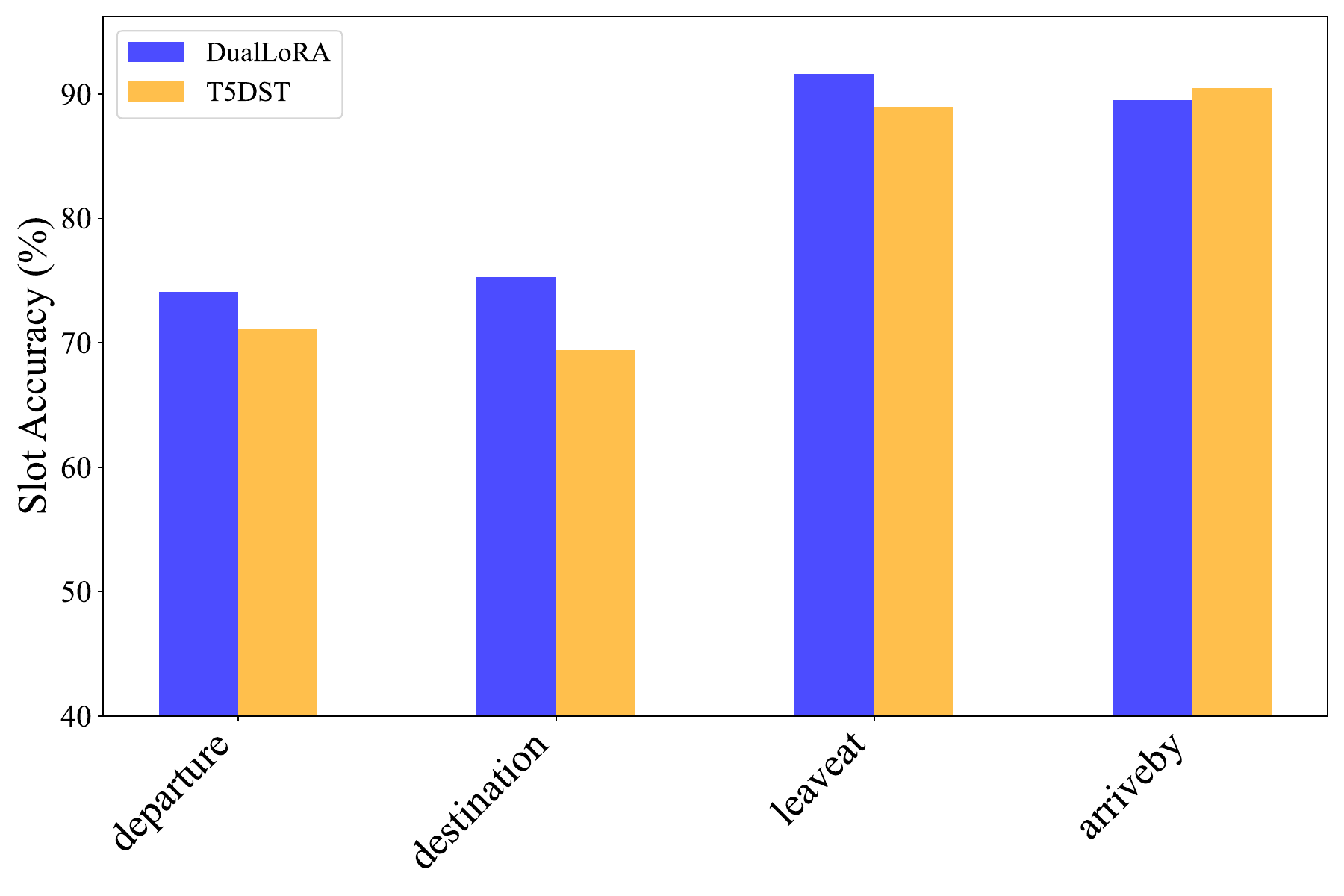}
    \caption{Slot Accuracy in taxi domain on MultiWOZ dataset.}
    \label{fig: taxi_acc}
\end{figure}

\begin{figure}[t!]
    \centering
    \includegraphics[width=0.49\textwidth]{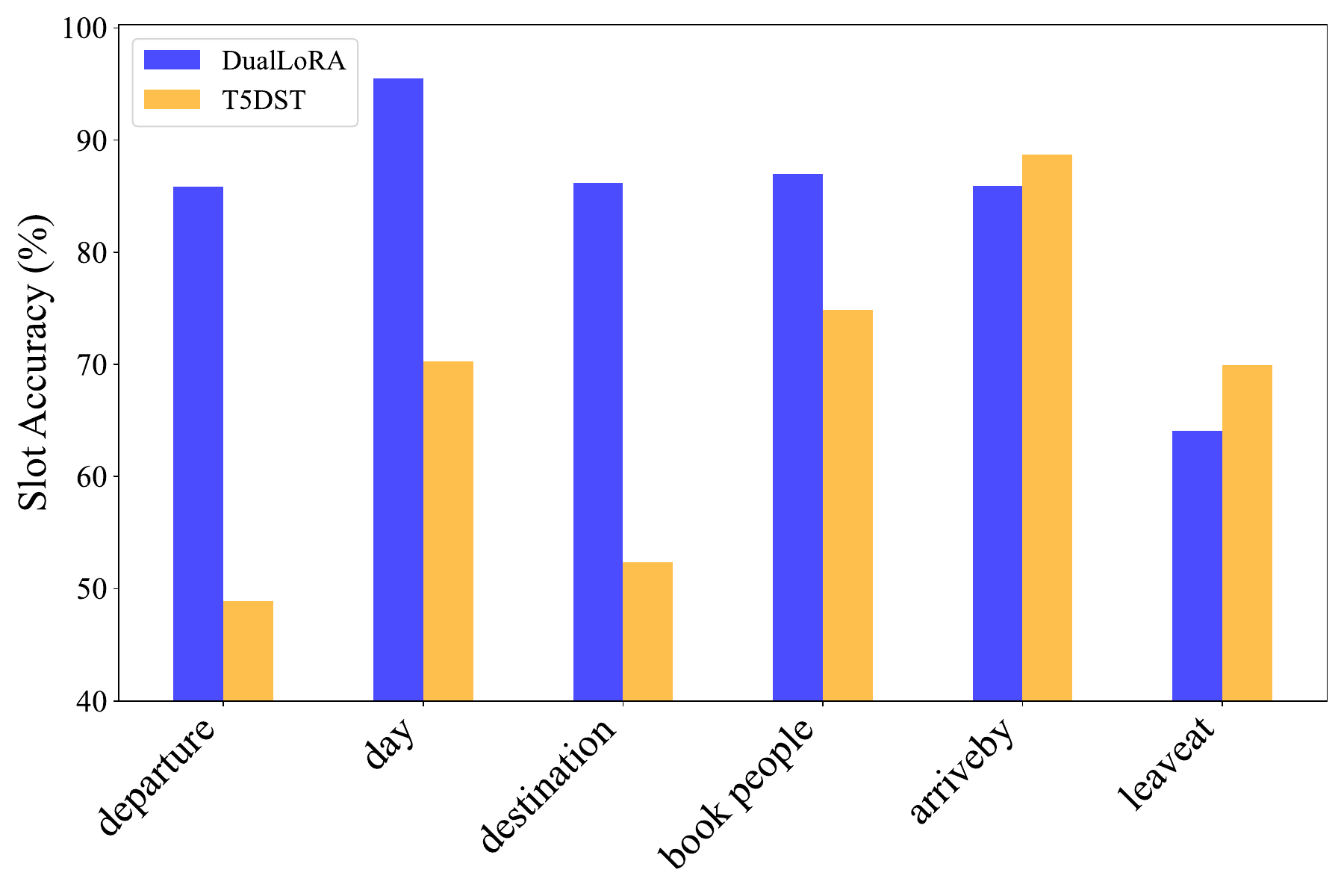}
    \caption{Slot Accuracy in train domain on MultiWOZ dataset.}
    \label{fig: train_acc}
\end{figure}

To delve into the slot accuracy performance of our model in each domain, we conducted experiments on each slot in every domain. The experimental results are presented in Figures \ref{fig: attraction_acc}, \ref{fig: hotel_acc}, \ref{fig: restaurant_acc}, \ref{fig: taxi_acc} and \ref{fig: train_acc}. We compared our model with T5DST and observed that in most domains and for most slots, DualLoRA performs better than T5DST, likely due to DualLoRA's deeper understanding of prompts.

\section{Impact of Different Prompt Input}
\begin{figure}[htb!]
    \centering
    \includegraphics[width=0.49\textwidth]{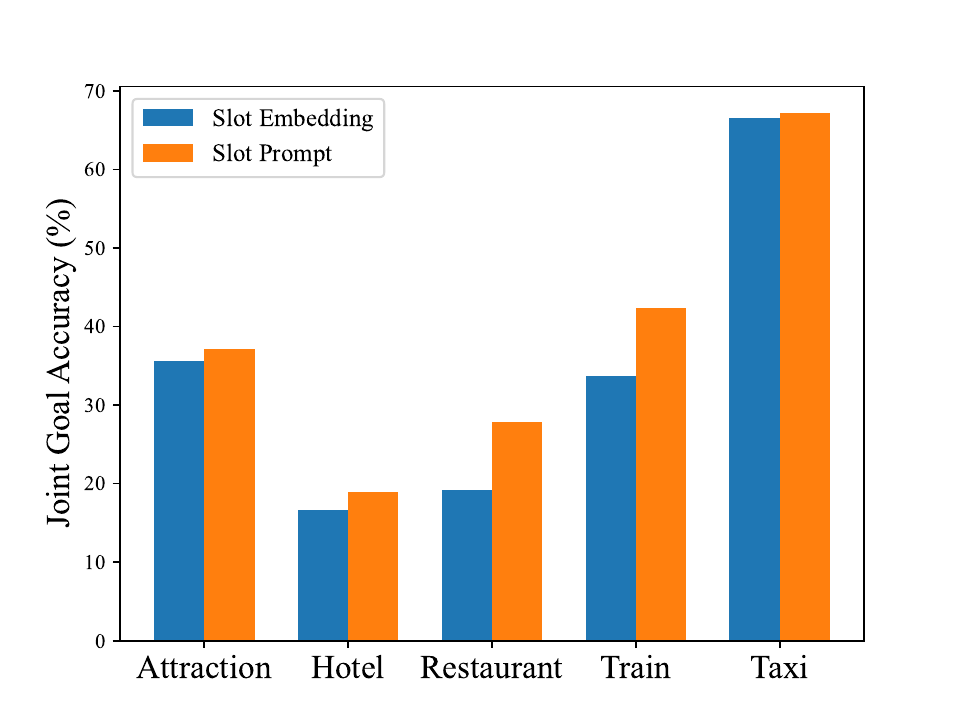}
    \caption{Performance with different prompt input on MultiWOZ dataset.}
    \label{fig: slot prompt}
\end{figure}

We conducted experiments using various inputs to explore the impact of different prompt inputs on the model's performance. We conducted experiments using regular slot embedding and slot prompts proposed by \citet{aksu_prompter_2023}, and the results are shown in Figure \ref{fig: slot prompt}. As indicated, slot prompt outperforms slot embedding in all domains. We analyze this because the differences in slot embedding within the same domain are minimal, and the model cannot learn effective prompts from them. In addition, during cross-domain zero-shot transfer, the differences in slot embedding between different domains are substantial, leading to the model's inability to utilize prompts in the target domain during cross-domain transfer effectively.

\section{Examples of Case Studies}
\begin{figure*}[t!]
    \centering
    \includegraphics[width=0.98\textwidth]{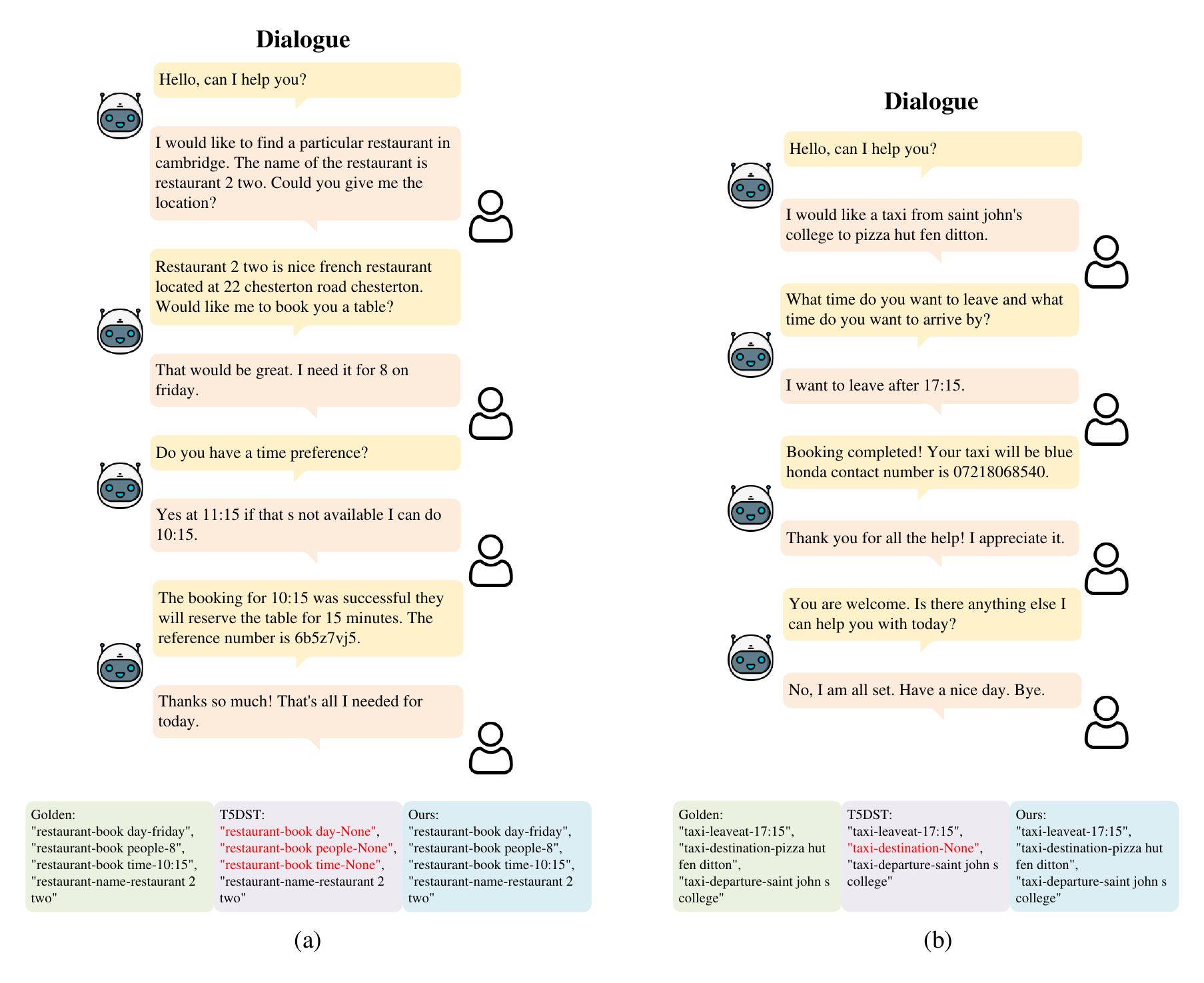}
    \caption{Example of zero-shot dialogue state tracking based on different models.}
    \label{fig: case 1}
\end{figure*}

\begin{figure*}[t!]
    \centering
    \includegraphics[width=0.98\textwidth]{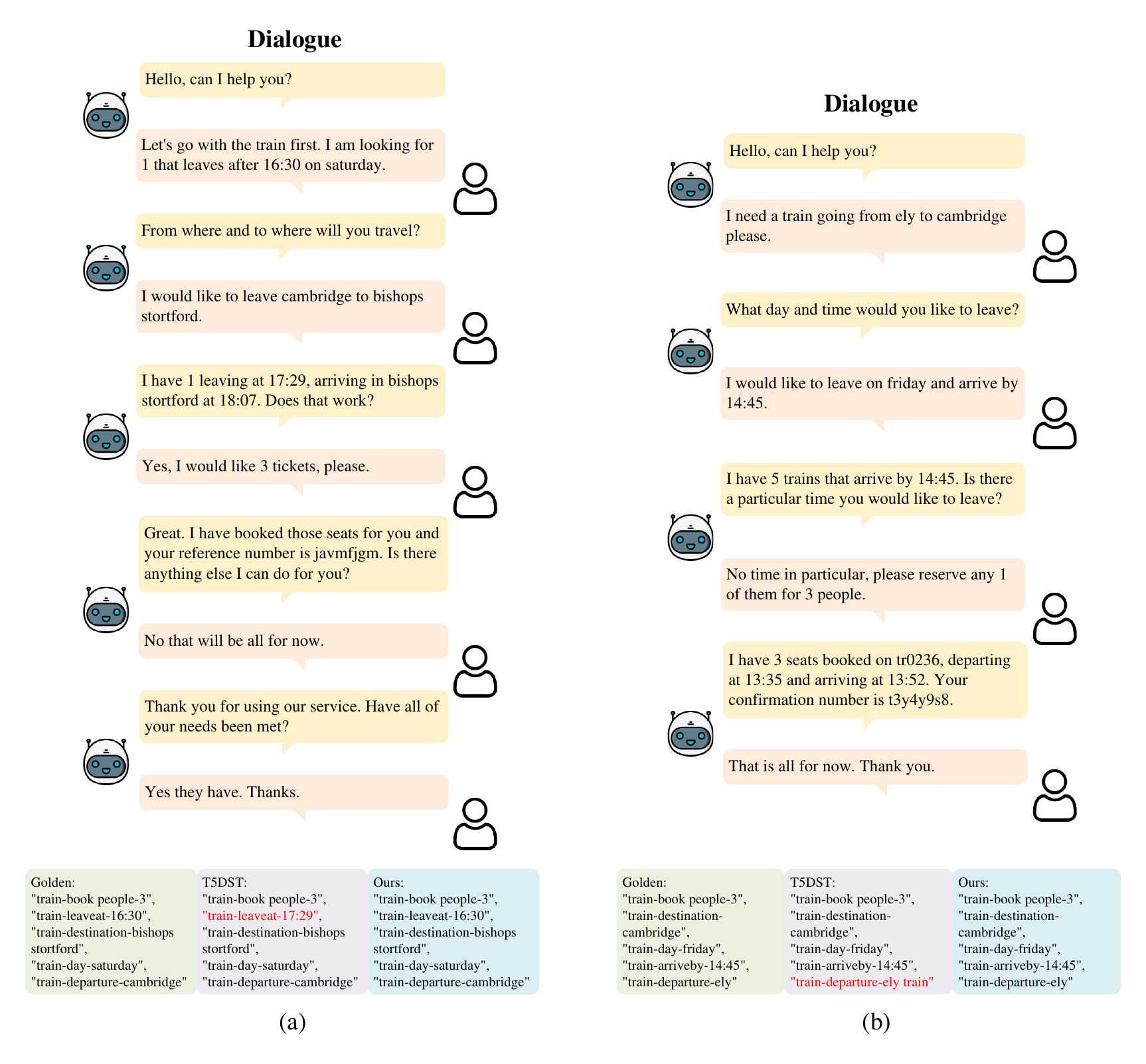}
    \caption{Example of zero-shot dialogue state tracking based on different models.}
    \label{fig: case 2}
\end{figure*}

\begin{figure*}[t!]
    \centering
    \includegraphics[width=0.98\textwidth]{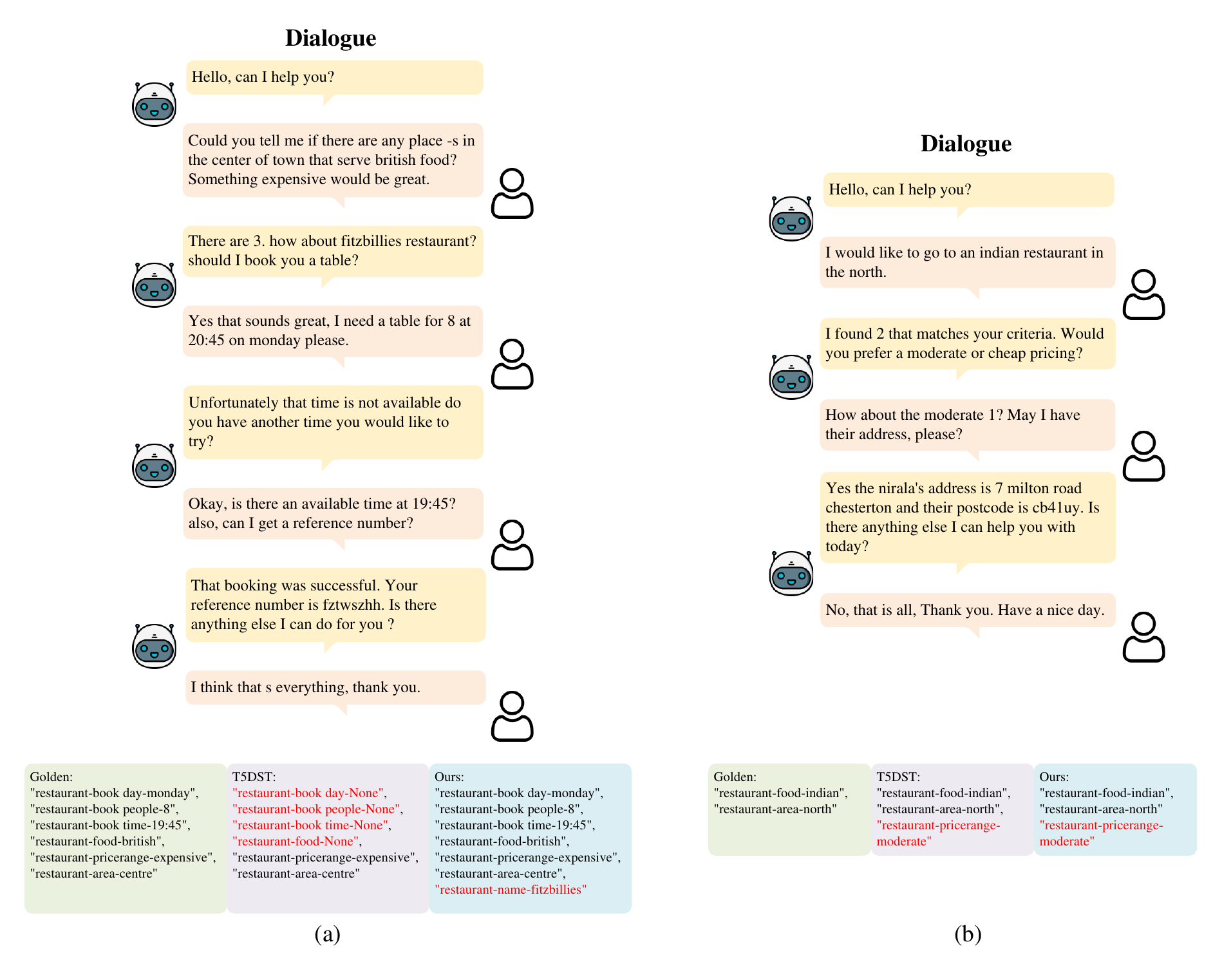}
    \caption{Example of zero-shot dialogue state tracking based on different models.}
    \label{fig: case 3}
\end{figure*}

This section presents a more detailed analysis of case studies. We categorize potential errors into three cases: under-predictions, incorrect-predictions, and over-predictions, as illustrated in Figures \ref{fig: case 1}, \ref{fig: case 2}, and \ref{fig: case 3}, respectively. The sentences marked in red indicate wrong predictions.

As seen in Figure \ref{fig: case 1}, our model enhances performance by leveraging prompts effectively, while T5DST's understanding of prompts remains superficial. Therefore, as shown in the figure, our model outperforms T5DST in terms of under-predictions in the restaurant and taxi domains. 

As shown in Figure \ref{fig: case 2}, the model's overfitting in the original domain may lead to incorrect-predictions during cross-domain zero-shot transfer. Additionally, the misunderstanding of slot values also contributes to incorrect-predictions. Therefore, our model, compared to T5DST, is effective in avoiding incorrect-predictions during cross-domain zero-shot transfer.

As shown in Figure \ref{fig: case 3}, our model makes errors in over-prediction. However, the predicted slot values by our model can still be extracted from the conversation. For instance, as stated in figure (a), \textit{"How about fitzbillies restaurant?"} and \textit{"Yes that sounds great."} suggests our model can extract the slot value for `name-fitzbillies` and as stated in figure (b), \textit{"How about the moderate 1?"} suggests our model can extract the slot value for `pricerange-moderate`, but it does not match the ground-truth slot value. We attribute this discrepancy to potential failures in data annotation.

\begin{figure*}[t!]
    \centering
    \includegraphics[width=0.98\textwidth]{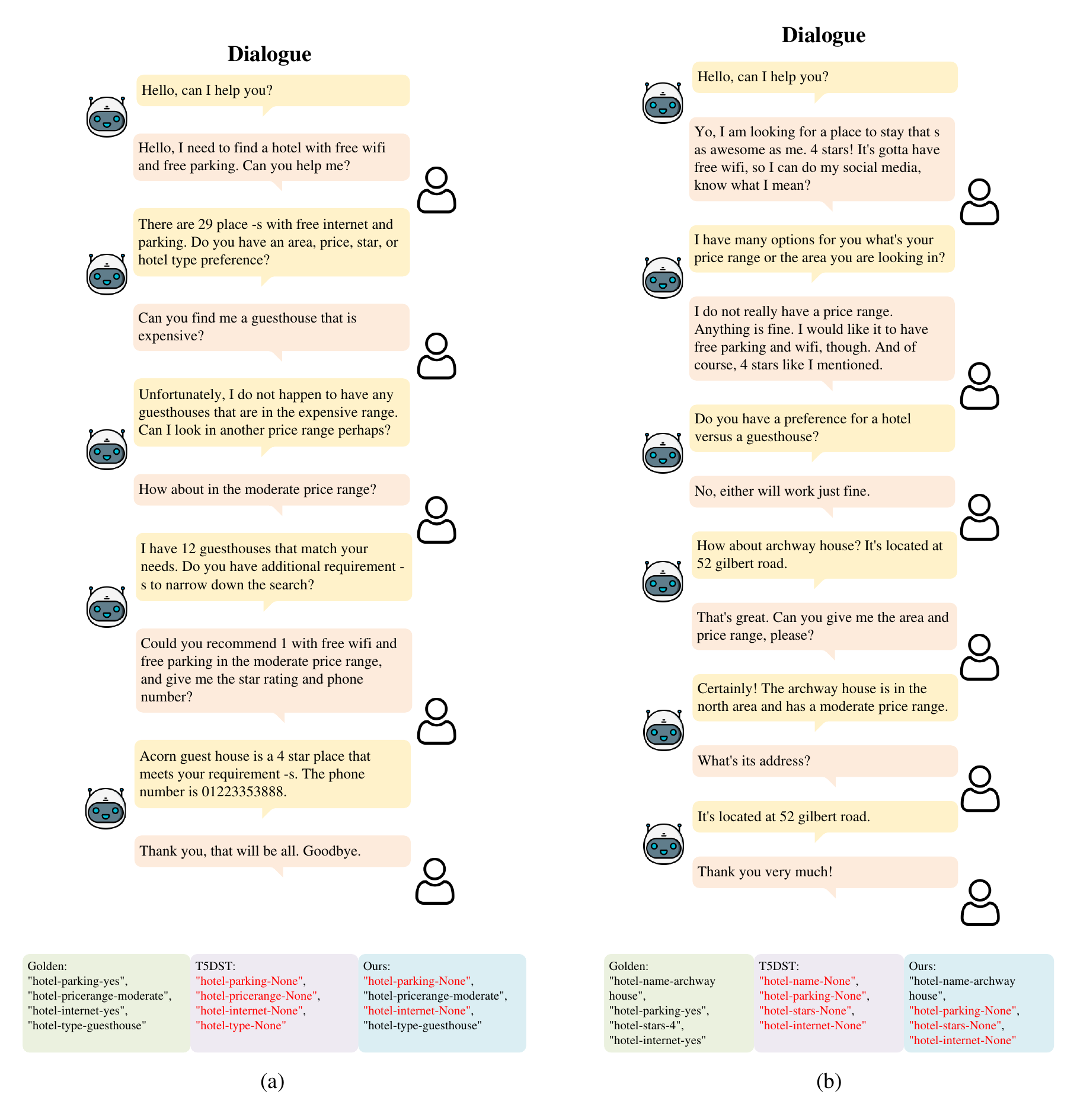}
    \caption{Example of zero-shot dialogue state tracking based on different models.}
    \label{fig: case 4}
\end{figure*}

In addition, our model has some limitations, as shown in Figure \ref{fig: case 4}. Our model fails to make predictions for some slots in the hotel domain, such as parking, internet, and stars. This might be related to the base model we adopted, T5-small. Due to the existence of some slots in the hotel domain that cannot be easily related to other domains, our model struggles to accurately track the values of these slots.

\section{Examples of Attention Map in Different Domains}
\begin{figure}[htb!]
    \centering
    \includegraphics[width=0.49\textwidth]{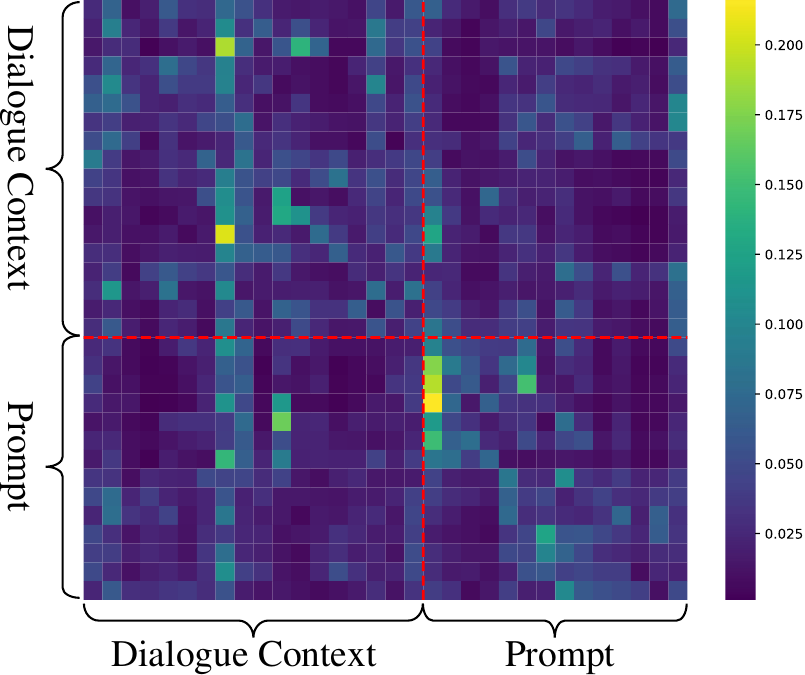}
    \caption{An attention map of DualLoRA between dialogue context and prompt in attraction domain.}
    \label{fig: attraction_attn}
\end{figure}

\begin{figure}[htb!]
    \centering
    \includegraphics[width=0.49\textwidth]{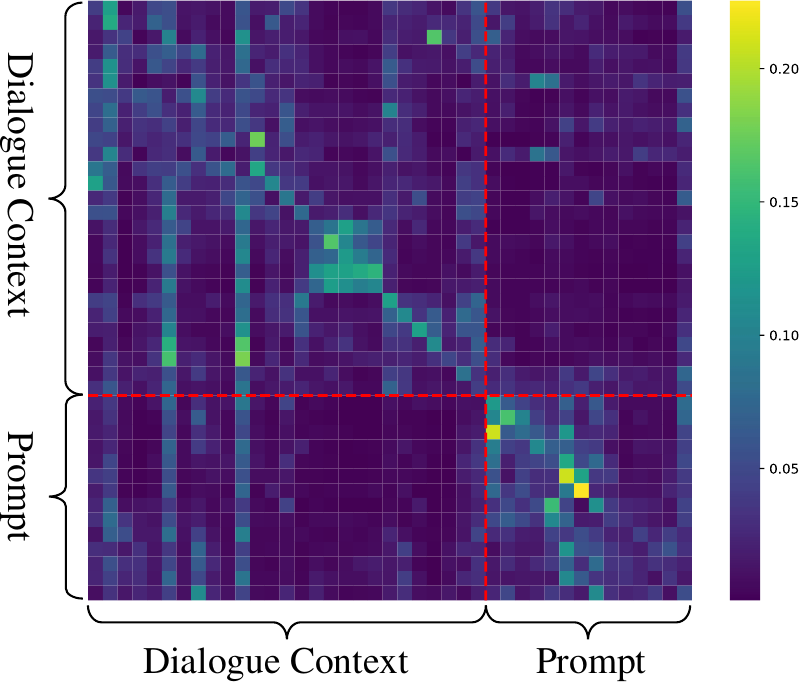}
    \caption{An attention map of DualLoRA between dialogue context and prompt in hotel domain.}
    \label{fig: hotel_attn}
\end{figure}

\begin{figure}[htb!]
    \centering
    \includegraphics[width=0.49\textwidth]{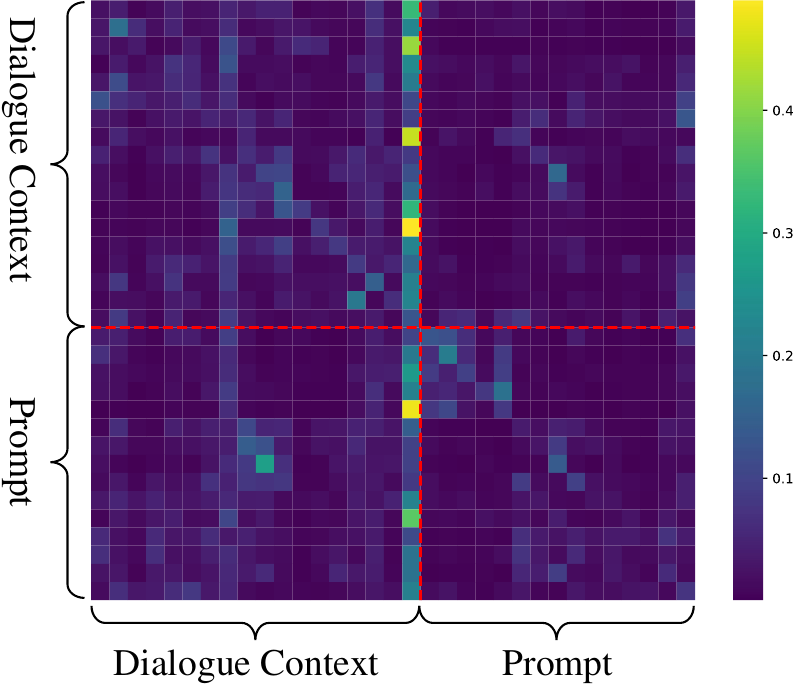}
    \caption{An attention map of DualLoRA between dialogue context and prompt in restaurant domain.}
    \label{fig: restaurant_attn}
\end{figure}

\begin{figure}[htb!]
    \centering
    \includegraphics[width=0.49\textwidth]{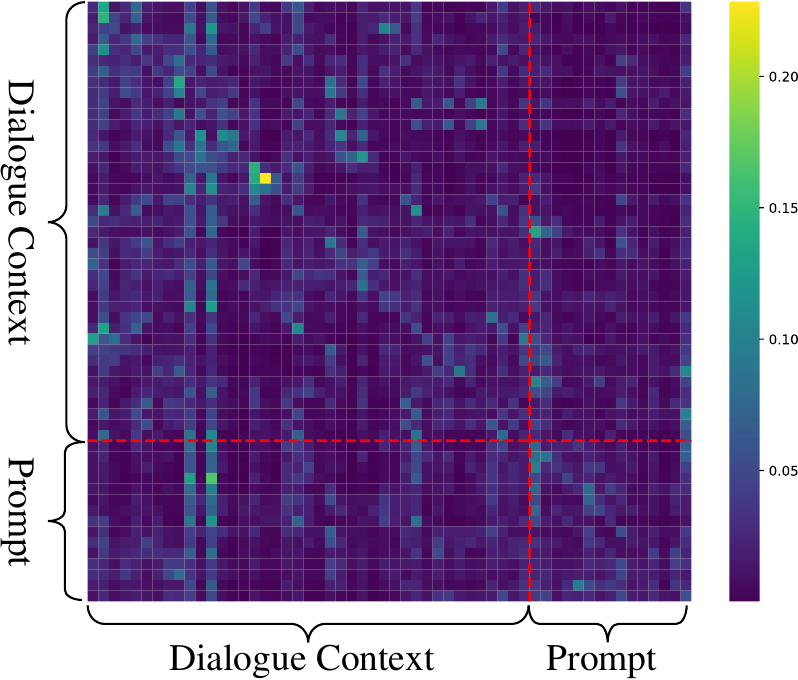}
    \caption{An attention map of DualLoRA between dialogue context and prompt in train domain.}
    \label{fig: train_attn}
\end{figure}

\begin{figure}[htb!]
    \centering
    \includegraphics[width=0.49\textwidth]{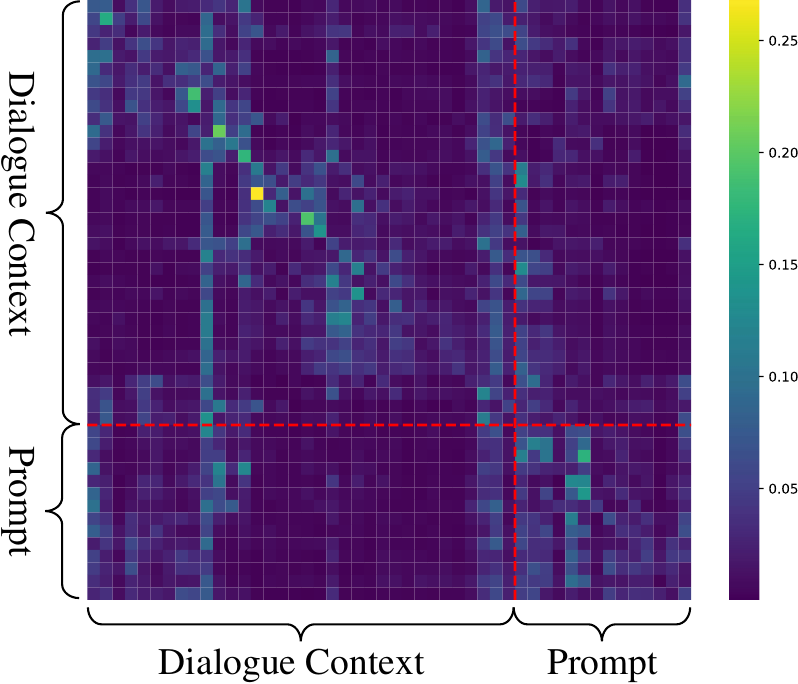}
    \caption{An attention map of DualLoRA between dialogue context and prompt in taxi domain.}
    \label{fig: taxi_attn}
\end{figure}

We present the visualization of attention in different domains after training with DualLoRA, as shown in Figures \ref{fig: attraction_attn}, \ref{fig: hotel_attn}, \ref{fig: restaurant_attn}, \ref{fig: train_attn} and \ref{fig: taxi_attn}, which can be observed that our model effectively focuses on the prompt part after training, rather than solely on the main diagonal.

\end{document}